%% file: 2025_acl_advcal.tex
\pdfoutput=1

\documentclass[11pt]{article}

\usepackage{style/acl_arr}

\newcommand*\samethanks[1][\value{footnote}]{\footnotemark[#1]}
\usepackage{times}
\usepackage{latexsym}
\usepackage[T1]{fontenc}
\usepackage[utf8]{inputenc} 
\usepackage{microtype}
\usepackage{graphicx}
\usepackage{amsmath, amssymb, amsfonts, amsthm}
\usepackage{booktabs}
\usepackage{siunitx}
\usepackage{algorithm}
\usepackage[noend]{algpseudocode}
\usepackage{enumitem}
\setlist{nosep}
\usepackage{multirow}
\usepackage{tabularx}
\usepackage{float}
\usepackage{makecell}
\usepackage{bbm}
\usepackage{bm}
\PassOptionsToPackage{dvipsnames,svgnames,x11names,table,xcdraw}{xcolor}
\usepackage{xcolor}
\usepackage[most]{tcolorbox}
\usepackage{subcaption}
\usepackage{xifthen} 
\usepackage{fontawesome5}

\definecolor{Violet}{RGB}{138, 43, 226}  
\definecolor{RoyalPurple}{RGB}{120, 81, 169}  
\definecolor{Purple}{RGB}{128, 0, 128}  
\definecolor{WildStrawberry}{RGB}{255, 67, 164} 
\definecolor{RedOrange}{RGB}{255, 83, 73}  
\definecolor{YellowOrange}{RGB}{255, 174, 66}

\newcommand{\buzz}[0]{buzz}

\newcommand{\roundbox}[2]{%
    \tcbox[colback=#1!20, colframe=#1!50!black, 
            boxrule=0.5pt, 
           left=3pt, right=3pt, top=1pt, bottom=1pt, 
           on line, boxsep=0pt]%
    {\faBell\ \textbf{#2}}%
}

\newcommand{\rankbox}[1]{%
    \def\colorcode{}%
    \ifthenelse{\equal{#1}{1}}{\def\colorcode{RoyalPurple}}{}
    \ifthenelse{\equal{#1}{2}}{\def\colorcode{violet}}{}
    \ifthenelse{\equal{#1}{3}}{\def\colorcode{purple}}{}
    \ifthenelse{\equal{#1}{4}}{\def\colorcode{WildStrawberry}}{}
    \ifthenelse{\equal{#1}{5}}{\def\colorcode{RedOrange}}{}
    \ifthenelse{\equal{#1}{6}}{\def\colorcode{YellowOrange}}{}
    \ifthenelse{\isempty{\colorcode}}{\def\colorcode{Apricot}}{}
    \raisebox{-0.5ex}{
    \begin{tcolorbox}[
        width=0.4cm,
        height=0.4cm,
        colback=\colorcode,
        colframe=\colorcode,
        arc=0pt,
        boxrule=0.5mm,
        valign=center,
        halign=center,
        nobeforeafter,
        boxsep=-3.5mm
    ]
        \textcolor{white}{\textbf{#1}}
    \end{tcolorbox}
    }
}
\renewcommand{\paragraph}[1]{\vspace{1ex}\noindent\textbf{#1.}}
\newcommand{\name}{\textsc{GRACE}}
\newcommand{\metric}{\textsc{CalScore}}

\newcommand{\buzzpoints}{buzzpoints}
\newcommand{\E}[1]{\mathbb{E}_t\left[#1\right]}

\definecolor{ForestGreen}{RGB}{34,139,34}

\newcommand{\qbquestion}[2]{%
  \noindent\begingroup
  \setlength{\parskip}{4pt} 
  \setlength{\parindent}{0pt} 
  \small
  \textbf{Q:} #1\par
  \textbf{ANSWER:} #2\par
  \endgroup
}

\graphicspath{ {figures/}{auto_fig/} }

\newif\ifcomment\commenttrue
\input{style/preamble}

\title{
    \name{}: A Granular Benchmark for Evaluating Model Calibration Against Human Calibration}

\author{Yoo Yeon Sung$^{1}$\thanks{\; Equal contribution.} \hspace{0.1cm}
Eve Fleisig$^{2}$\samethanks \hspace{0.2cm} 
\textbf{Yu Hou}$^{1}$ \hspace{0.1cm} 
\textbf{Ishan Upadhyay}$^{3}$ \hspace{0.1cm} 
\textbf{Jordan Boyd-Graber}$^{1}$\\
  $^{1}$University of Maryland \hspace{0.5cm}
  $^{2}$UC Berkeley \hspace{0.5cm} \hspace{0.5cm}
  $^{3}$IIT Bombay \hspace{0.5cm}
}

\date{}

\begin{document}
\maketitle
\begin{abstract}
\input{2025_acl_advcal/sections/00-abstract}

\end{abstract}
\input{2025_acl_advcal/sections/10-intro}

\input{2025_acl_advcal/sections/20-preliminaries}

\input{2025_acl_advcal/sections/30-dataset-development}
\input{2025_acl_advcal/sections/50-advcal-systemmetric}
\input{2025_acl_advcal/sections/60-calibration-evaluation}
\input{2025_acl_advcal/sections/62-metric-analysis}

\input{2025_acl_advcal/sections/70-conclusion}
\input{2025_acl_advcal/sections/80-limitation}

\bibliography{bib/yy,bib/jbg}
\bibliographystyle{style/acl_natbib}

\clearpage
\appendix
\input{2025_acl_advcal/sections/appendix}

\end{document}

%% file: style/preamble.tex
\usepackage[a-1b]{pdfx}

\usepackage{framed}
\usepackage{mdwlist}
\usepackage{siunitx}
\usepackage{latexsym}
\usepackage{colortbl}
\usepackage{xcolor}
\usepackage{nicefrac}
\usepackage{booktabs}
\usepackage{fnpct}
\usepackage{amsfonts}
\usepackage[T1]{fontenc}
\usepackage{bold-extra}
\usepackage{amsmath}
\usepackage{amssymb}
\usepackage{bm}
\usepackage{graphicx}
\usepackage{mathtools}
\usepackage{microtype}
\usepackage{multirow}
\usepackage{multicol}
\usepackage{threeparttable}
\usepackage{xpatch}
\usepackage{latexsym,comment}
\usepackage[normalem]{ulem}

\newcommand*{\missingreference}{{\Huge \colorbox{red}{?reference?}}}
\newcommand*{\missingcitation}{{\Huge \colorbox{red}{?citation?}}}

\makeatletter
\xpatchcmd{\@setref}{\bfseries}{\missingreference}{}{}
\def\@citex[#1]#2{\leavevmode
    \let\@citea\@empty
    \@cite{\@for\@citeb:=#2\do
        {\@citea\def\@citea{,\penalty\@m\ }%
            \edef\@citeb{\expandafter\@firstofone\@citeb\@empty}%
            \if@filesw\immediate\write\@auxout{\string\citation{\@citeb}}\fi
            \@ifundefined{b@\@citeb}{\hbox{\reset@font\missingcitation}%
                \G@refundefinedtrue
                \@latex@warning
                {Citation `\@citeb' on page \thepage \space undefined}}%
            {\@cite@ofmt{\csname b@\@citeb\endcsname}}}}{#1}}
\makeatother

\newcommand{\gem}[1]{\mbox{\textsc{gem}}}
\newcommand{\abr}[1]{\textsc{#1}}

\newcommand{\g}{\, | \,}

\renewenvironment{quote}
{\list{}{\rightmargin\leftmargin}%
    \item\relax\small\ignorespaces}
{\unskip\unskip\endlist}

\DeclareUnicodeCharacter{03B2}{ }
\DeclareUnicodeCharacter{2212}{ }

\newcommand{\hidetext}[1]{}
\newcommand{\ignore}[1]{}

\ifcomment
    \newcommand{\pinaforecomment}[3]{\colorbox{#1}{\parbox{.8\linewidth}{#2: #3}}}

    \newcommand{\prtodo}[1]{\pinaforecomment{lightblue}{pr}{#1}}
    \newcommand{\prtodoi}[1]{\pinaforecomment{lightblue}{pr}{#1}}
\else
    \newcommand{\pinaforecomment}[3]{}
    \newcommand{\prtodo}[1]{}
    \newcommand{\prtodoi}[1]{}
\fi

\newcommand{\smallurl}[1]{ \begin{tiny}\url{#1}\end{tiny}}

\definecolor{lightblue}{HTML}{3cc7ea}
\definecolor{CUgold}{HTML}{CFB87C}
\definecolor{grey}{rgb}{0.95,0.95,0.95}
\definecolor{ceil}{rgb}{0.57, 0.63, 0.81}
\definecolor{UMDred}{HTML}{ed1c24}
\definecolor{UMDyellow}{HTML}{ffc20e}


%% file: 2025_acl_advcal/sections/00-abstract.tex
Language models are often miscalibrated, leading to confidently incorrect answers.
We introduce \name{}, a benchmark for language model
calibration that incorporates comparison with human calibration.
\name{} consists of question-answer pairs, in which each question contains a series of clues that gradually become easier, all leading to the same answer; models must answer correctly as early as possible as the clues are revealed.
This setting permits granular measurement of model calibration based on how early, accurately, and confidently a model answers.
After collecting these questions, we host live human vs. model competitions to gather 1,749 data points on human and model teams' timing, accuracy, and confidence.
We propose a metric, \metric{}, that uses \name{} to analyze model calibration errors and identify types of model miscalibration that differ from human behavior. We find that although humans are less \emph{accurate} than models, humans are
generally better \emph{calibrated}. Since state-of-the-art models
struggle on \name{}, it effectively evaluates progress on improving model calibration.\footnote{Code and data:  
\url{https://github.com/yysung/advcalibration}}

%% file: 2025_acl_advcal/sections/10-intro.tex
\section{Introduction}\label{sec:intro}


Because language models are often miscalibrated, they are often confidently wrong~\citep{kaur2020interpreting}.
This mismatch between accuracy and confidence causes users to trust models more than they
should~\citep{caruana2019friends, deng2025development}, even over their own correct judgment~\citep{krause2023confidently,
  stengel2023calibrated,liu2024dellma}.
These issues are particularly severe when models are miscalibrated in ways that humans are not: users expect models to be at least as calibrated as humans, and when models are worse, users are often not prepared to address these errors. Thus, models should be \textit{at least} as calibrated as humans, making it especially crucial to identify when models commit calibration errors that humans do not. However, existing work on model calibration lacks comparison with human calibration.

\begin{figure*}[!t] 
    \centering
    \includegraphics[width=\linewidth]{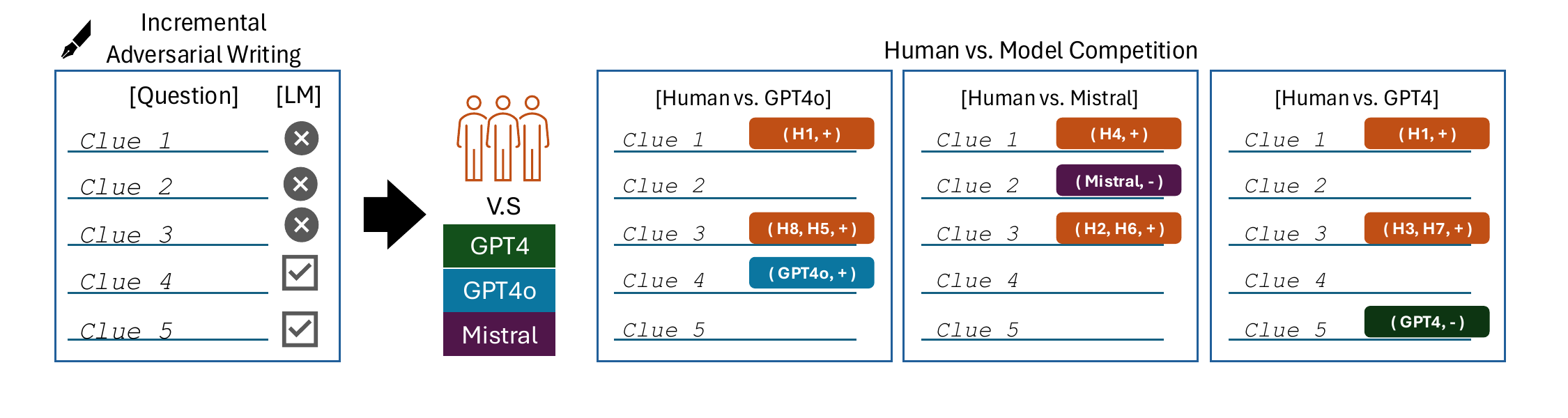}
    \caption{To create the \name{} dataset, expert question writers develop questions with multiple clues of decreasing difficulty via an interface that shows where weaker models struggle to answer the questions.
      These questions are used in human vs. model competitions where teams compete to be the first to interrupt the sequence of clues with a correct answer. We record when the human and model teams buzz in each question with their correctness (+) or incorrectness (-) (\textit{buzzpoints} \faBell). The dataset contains all buzzpoints throughout the competition. Then, \metric{} measures each model's human-grounded calibration performance (\S~\ref{sec:main-metric}).}
    \label{fig:dataset_pipe}
\end{figure*}

We thus introduce \name{}, a  \textbf{Gra}nular, Human-grounded Benchmark for Model \textbf{C}alibration \textbf{E}valuation.
Each instance allows \textbf{fine-grained calibration measurement} using
an incremental question-answering (\abr{qa}) framework.
%
%
Expert writers design \name{} questions, each consisting of at least five sentences of clues that gradually become easier.
%
%
%
To prevent models from being confused by ambiguities or false
presuppositions, we require that clues challenge models but remain
clear for humans.
%
This format measures model calibration with human performance as a reference point: models should give correct answers earlier and more confidently than humans, while minimizing confidently incorrect guesses (\S~\ref{sec:dataset}).

%
\name{} incorporates human responses from live \abr{qa} competitions we conduct. Unlike prior calibration evaluation methods that only allow model--model
calibration comparisons, our dataset thus allows direct
\textit{human--model} calibration comparison.
\textbf{\name{} is the first benchmark dataset designed to
  evaluate model calibration grounded in human needs}.

This unique dataset is the foundation for a new metric (\metric{}, \S~\ref{sec:main-metric}).
In contrast to other calibration evaluation methods that only calculate aggregate calibration over
the entire dataset, \name{} also facilitates per-instance evaluation, which
helps in identifying specific contexts where models are much worse than humans at avoiding confidently incorrect answers.

%
We find that \textbf{language models are more overconfident than humans in incorrect
  answers, and relatively underconfident in
   correct answers.} In contrast, humans tend to be highly
  confident---over 50\%---when correct (\S~\ref{sec:analysis-bayes}).
Models struggle with abstract descriptions---they are both
overconfident and inaccurate---but excel in retrieving facts
given unambiguous clues (\S~\ref{sec:qual_analysis}).
%
We conclude with a discussion of how \name{} and \metric{} can aid in the
creation of models that are more accurate and better calibrated.

%% file: 2025_acl_advcal/sections/20-preliminaries.tex
\section{Preliminaries}\label{sec:compare}











Drawing on prior work, \name{} consists of incremental questions with adversarial clues to effectively ground model calibration evaluation in human performance.
%



\paragraph{Incremental and adversarial QA}
Incremental questions contain multiple clues in decreasing order of 
difficulty; models must answer correctly as early as
possible.
\citet{boyd-graber-etal-2012-besting} and \citet{he-16} argue that
this setting is a natural test of calibration:
participants should buzz only when confident in their answers 
~\citep{ferrucci2012introduction}. Unlike
selective \abr{qa} for non-incremental
examples~\citep{kamath-etal-2020-selective}, incremental decision processes offer more detail, since each clue is a decision point for whether to answer or abstain, with 
more information available as clues are revealed.

In addition, unlike prior incremental \abr{qa} research, we use a human-in-the-loop adversarial authoring process to specifically
target calibration.
\citet{quizbowl2019} uses publicly available questions, which are too easy for modern models.
While \citet{wallace2019trick} used model--human collaboration for
incremental adversarial data, their questions are also insufficiently adversarial.\footnote{ GPT-4 has 80\% accuracy on \citet{wallace2019trick}'s TrickMe dataset after only 60\% of the clues (Appendix~\ref{app:trickme-comparison}). Model accuracy remains 2-4x higher on TrickMe than on \name{} as clues are revealed.} In addition, they did not account for model
confidence, treating buzzing as a binary outcome.

To develop questions that challenge models, expert writers use our interface (\S~\ref{sec:question-writing}), followed by expert editing to ensure well-posed
questions.
%
Our dataset creation process is motivated by \citet{kiela-etal-2021-dynabench,
  ilyas2019adversarial, engstrom2019adversarial}, who argue
that adversarial benchmarks must be clear for humans and challenge
models, ensuring that model errors are due to model limitations rather than ambiguous or low-quality questions~\citep{min2020ambigqa, yu-etal-2023-crepe}.
%

\textbf{Grounding to human calibration.}
Humans increasingly use \abr{ai} to help make decisions, but such assistance can be detrimental when the model is
miscalibrated~\citep{StengelEskin2024LACIELF} or fails to
abstain~\citep{khurana2024crowdcalibratorannotatordisagreementinform}.
This is particularly concerning when models are confidently wrong but humans do not know the correct answer, which our metric, \metric{}, especially penalizes.
Furthermore, modern models are poorly
calibrated to human linguistic variation, causing
\citet{ilia2024predict} to question the
reliability of expected calibration error (ECE).
Thus, \metric{} focuses on where models can help users by
considering how early \textit{humans} can answer the
questions.
\input{2025_acl_advcal/sections/interface_sample}

\textbf{Calibration evaluation.}
Language models tend to be overconfident in their predictions, which
can lead to undue trust or erode user confidence in language models
\citep{zhou-etal-2024-relying}.
%
Proposed methods to measure model calibration include using raw
probabilities \cite{xiong2024llmsexpressuncertaintyempirical}, separate confidence
predictors \cite{ulmer-etal-2024-calibrating}, verbalized confidence
scores \cite{tian-etal-2023-just,
  band2024linguisticcalibrationlongformgenerations}, or natural
language expressions of uncertainty \cite{StengelEskin2024LACIELF,
  zhou-etal-2023-navigating}.
%
%
Our dataset aids finer-grained versions of these approaches by permitting per-instance, human-grounded
calibration measurement.
We also extend on existing calibration metrics, such as
ECE~\citep{naeini2015obtaining} and Brier scores
\cite{brier1950verification} by introducing a metric for calibration
on incremental questions (Appendix~\ref{app:ece-brier}).

%% file: 2025_acl_advcal/sections/interface_sample.tex
 \begin{figure*}[!t] 
    \centering
    \includegraphics[width=\linewidth]{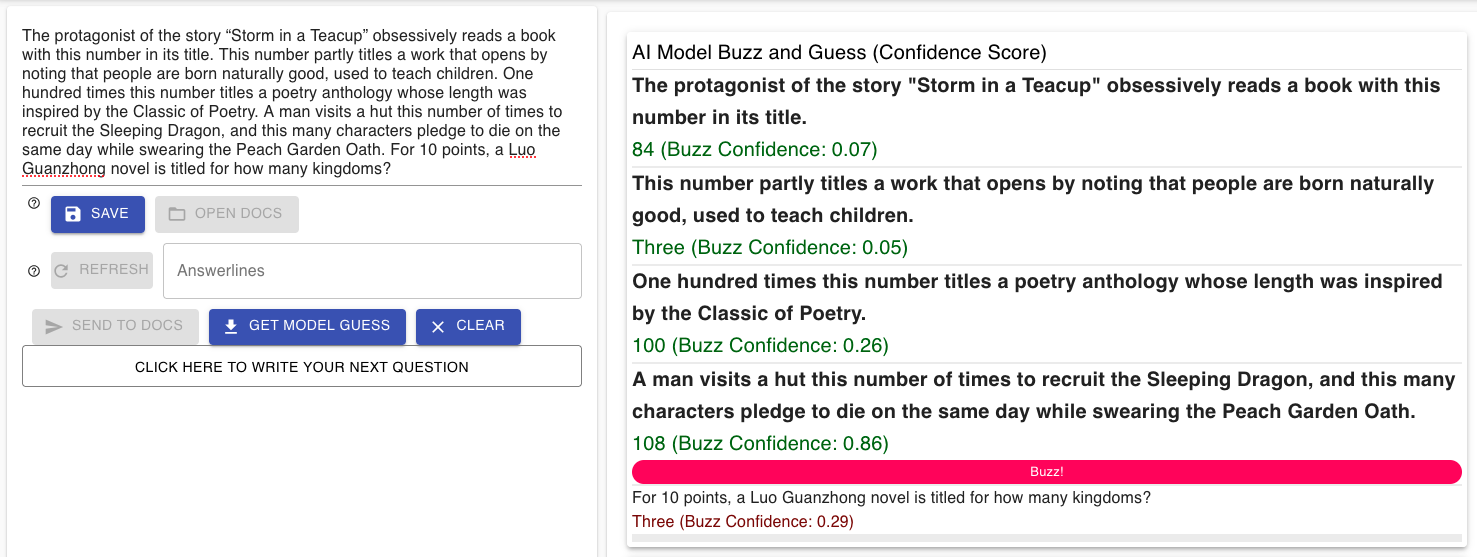}
    \caption{Example question on Chinese literature (with the answer of \underline{three}) being written in the interface. Writers compose questions in the left box. On the right, they see the model's guess and confidence after every sentence and the point at which the model would buzz in and attempt to answer. Writers learn which sentences make it harder for models to answer correctly and refine their questions to be sufficiently hard for models but still answerable by humans. This incremental, adversarial format permits granular calibration measurement.}
    \label{fig:sample_interface}
\end{figure*}

%% file: 2025_acl_advcal/sections/30-dataset-development.tex
\section{\name{}: Dataset Development}\label{sec:dataset}

%
To create our dataset, expert writers and editors first construct incremental,  adversarial, and rigorously quality-checked examples (\S~\ref{sec:question-writing}).
Then, we collect model guesses and confidences on these questions (\S~\ref{sec:buzzpoints}), and compare them against human performance in a live competition (\S~\ref{sec:tournament}).

\subsection{Question writing process}
\label{sec:question-writing}



\paragraph{Collecting QA pairs from expert writers}
We recruit experienced question writers and editors to ensure that questions are high-quality.
We hire six writers and ten editors to author the questions (qualifications in Appendix \ref{app:writer-qualifications}).
The questions contain 575--650 words\footnote{We refer to these as \textit{questions} although they are not grammatical questions, but rather sequences of sentences with clues uniquely identifying an answer (examples in Figures  \ref{fig:sample_interface} and \ref{fig:qb_vangogh}).} and cover content across six
categories.\footnote{Science, history, literature, and other (Appendix~\ref{app:dataset-design}).}
All questions are reviewed by the writer, category editor, and
head editor to check that clues are unambiguous and factually correct.
%

\paragraph{Interface setup}
To create incremental and adversarial questions, writers and editors
use a human-AI collaborative writing interface
(Figure~\ref{fig:sample_interface}).
Because these examples are meant to be incremental, we break the input
into sentences $\{s_1, s_2,\dots s_k\}$.
\abr{gpt}-3.5 provides a guess
$\{a_1, a_2,\dots a_k\}$ for
each sentence in addition to its confidence
$\{c_1, c_2\dots c_k\}$.
The interface also highlights when the model confidence would be high enough to buzz 
(Appendix~\ref{logistic}).

To ensure that questions remain incremental for models, we instruct
writers to write questions for which the model's guess is incorrect until the penultimate sentence or later.\footnote{The model’s confidence for \emph{correct} answers should remain low for all but the last two sentences; clues that trigger high-confidence, \emph{incorrect} model guesses are encouraged.}
As experienced question writers, they use their domain knowledge to ensure difficulty also decreases for humans, so that most humans can
answer correctly by the end.
%
Writers
dynamically interact with the models to refine their questions
~\citep{you-lowd-2022-towards}.
For example, the second line in Figure~\ref{fig:sample_interface}
was originally ``This number of characters appear in the name of a
Chinese classic,'' which the model answered correctly. Instead, the
editor revises the first line of the text, which
fools the models while allowing humans to answer correctly.
The final dataset consists of 243 QA pairs, with a total of 1,236 sentences of clues. Each sentence uniquely points to the answer, making it usable as a standalone QA pair.\footnote{For purposes of our analysis, a \textit{clue} is a substring of the full question, averaging 13 words or 35 characters, incrementally extending from the beginning. Clues are split at whitespace boundaries and may contain multiple pieces of information about the answer.}



\subsection{Collecting human--model buzzpoints}
\label{sec:buzzpoints}

The questions described above are designed to be read aloud and interrupted. In the competition, teams compete by buzzing to interrupt and answer, with this timing referred to as \buzzpoints{} \faBell~(Figure~\ref{fig:qb_vangogh}).
However, modern \abr{llm}s do not operate this way: they generate an output given an input.
Thus, we first extract guesses from models and humans \textit{offline} to assess teams on the same questions (\S \ref{sec:offline}). We then compute the model buzzpoints for each clue. Finally, using these precomputed model buzzpoints, we host live human–computer competitions to collect real-time human buzzpoints (\S \ref{sec:tournament}). 
%

\begin{figure}[t!]
    \centering
    \begin{minipage}{0.48\textwidth}
    \hrule\vspace{5pt}
    \qbquestion{In a reference to an object notably missing from one of these
    works, Diemut Strebe used genetic samples from a man's
    great-great-grandnephew to clone a certain feature. In one of these
    works, Utagawa Togokuni's Geishas in a Landscape
    \roundbox{red}{GPT-4o: ``Rodin statues''} hangs on a yellow wall
    behind a man in a fur-brimmed hat. The backside of The Potato Peeler
    includes one of these works featuring a man in a straw hat. One work
    shows a man in a light-blue green suit against a light-green-blue
    swirling background, and another dedicated to Gauguin shows subject
    with cropped hair and a red beard. For 10 points, a Dutch artist
    \textbf{\roundbox{ForestGreen}{H1: ``Van Gogh self-portraits''}}
    painted what portraits of himself with a bandaged ear?}
    {\underline{self-portraits of Vincent Van Gogh}}
    \vspace{5pt}\hrule
    \end{minipage}
    \caption{While GPT-4o buzzes too early with an \roundbox{red}{incorrect answer}, losing 5 points, the human team (H1) buzzes later with a \roundbox{ForestGreen}{correct answer}, earning 10 points. Both teams must balance accuracy and speed; here, \texttt{GPT-4o} shows poorer calibration than \texttt{H1}.}
    \label{fig:qb_vangogh}
\end{figure}

\subsubsection{Offline human and model buzzpoints}\label{sec:offline}
\textbf{Model guesses and confidence.}
%
We first break each question into clues.
We then retrieve a model's guess given the first $n$ clues with a prompt using a \abr{tf-idf} retriever to select
similar question-answer pairs from \abr{qa} datasets (Appendix~\ref{sec:guesser}). 
To determine if model guesses were correct, our post-processing uses both transformer-based answer equivalence (PEDANT,~\citealp{li-etal-2024-pedants}) and manual verification by dataset editors.
We store the resulting guesses and two forms of confidence from LLMs,
token logits\footnote{\name{} answers are short, typically 3-4 words long, making token logits a reliable measure of confidence.} and verbalized confidence.
Logit-based confidence reflects the average of the exponentials of the
token logit probabilities, while verbalized confidence prompts models to directly express confidence in the output tokens
(Appendix~\ref{sec:confidence_elicitation}). 


\textbf{Precomputed model \buzzpoints{}.}
While our metric, \metric{} (\S~\ref{sec:baseline-metric}), uses the raw confidence values from continuous probabilities, our
human--computer competitions (\S~\ref{sec:tournament}) require binarized confidence to indicate when the model buzzes.
%
For each model, we set a threshold based on human gameplay data on
preexisting non-adversarial questions~\citep{pmlr-v48-he16} (Appendix~\ref{app:buzzpoint-threshold}).
This threshold is chosen to maximize the probability of the model
buzzing correctly before the average trivia player as estimated by the
\textit{expected wins} metric from \citet{rodriguez2019quizbowl}.
When the logit score exceeds the threshold, the model buzzes
in, marking its buzzpoint (Appendix \ref{app:buzzpoint-algo}).

\textbf{Offline human guesses.}
\label{sec:survey}
To compute humans' raw accuracy, independent of confidence (\S~\ref{sec:analysis-bayes}),
we survey fifteen players on 35–40 held-out \name{} questions.
Like the models, players view clues, submit their guess after each clue, and indicate whether they would \buzz{} at that point.
However, this data collection format is time-consuming and tedious (one player called it ``remarkably hard''), potentially reducing player engagement and response quality. Instead, we collect human calibration data through a fast-paced trivia tournament.


%
%

\subsubsection{Human buzzpoints, \textit{live} competition}
\label{sec:tournament}

\name{} records human and computer guess correctness
on interruptible questions designed to challenge model calibration.
A human moderator reads each question to both teams (a model and a team
of humans).
Teams compete by buzzing to interrupt and answer.
Model \buzzpoints{} are computed in advance (\S~\ref{sec:offline}). When the model's confidence exceeds the threshold, the reading stops with a buzz sound, and the model's guess is announced.

\textbf{Human \buzzpoints{}.} 
In contrast, human buzzpoints are recorded in real time when the moderator is interrupted.
Players press a physical
buzzer when confident in an answer, and the moderator verifies if the answer is correct.
We log the timing of human teams'
buzzes and answer correctness.

If a team answers incorrectly, the moderator continues reading until the other team buzzes in.
Because earlier clues are harder, more skilled teams tend to buzz
earlier, while less skilled teams wait until near the end. Thus, teams must
be knowledgeable \emph{and} well-calibrated to buzz optimally.

\textbf{Human players in live competitions.}
Our three competitions consist of a total of 93 matches involving 17 human trivia teams and three LLMs (GPT-4o, GPT-4, and Mistral-7b-Instruct). Of these, 55 are human vs. model matches, while 38 are human vs. human matches. For the matches, the 243 QA pairs are divided into 12 sets of 20,
stratified by category, with three questions for tiebreakers.

Hosting real-time competitions with human players provides several benefits: (1) direct comparison of confidence calibration between humans and models on the same questions, (2) recruiting experienced players skilled in calibrating their answers,\footnote{After the competitions, we survey players about their experience levels and individual strengths (Appendix~\ref{app:player_specifics}).} and (3) validating that questions are human-answerable and unambiguous, as an additional quality check for the dataset.

%





%% file: 2025_acl_advcal/sections/50-advcal-systemmetric.tex





\section{Human-Grounded Calibration Evaluation}\label{sec:main-metric}
To compare model and human calibration, we analyze response correctness and buzz decisions. Then, we introduce a baseline metric, \metric{}. Unlike traditional calibration metrics, \metric{} facilitates per-instance calibration analysis, which lets us identify specific questions on which models are especially miscalibrated. In addition, it factors in human performance on the same question, to penalize cases in which models are confidently incorrect when humans are uncertain (\S~\ref{sec:baseline-metric}).



\subsection{Human-grounded metric: \metric{}}
\label{sec:baseline-metric}
\metric{} evaluates model calibration error while incorporating human buzzpoints. 
This adjustment reflects the structure of the competition---models must be confidently correct before humans know the answer. The adjustment also places higher weight on instances where model errors are more likely to mislead users---if a model is confidently incorrect when humans are still uncertain, humans are less likely to recognize and override the error (\S~\ref{sec:analysis-metric}).

Using the live competition data, we track the proportion of humans answering correctly up to a specific clue so that the metric applies higher penalties and rewards for earlier (harder) clues. 
To measure the expected probability of a team buzzing correctly on a given question, we consider teams' buzzes at each clue \(t\) of question \(q\). We define \( h_t \) as the cumulative probability of a human team correctly buzzing up to clue $t$, calculated as the number of correct buzzes by human teams up to \(t\) divided by total buzzes by human teams up to \(t\). For model responses, \( g_t \) indicates the correctness of a model's guess at clue \(t\) (1 if correct, -1 if not), and \( c_t \) indicates the model's confidence in its guess.

\subsection{Unadjusted model calibration error}
The normalized expectation $r (\E{g_t c_t})$, calculated over clues in a single question, measures calibration as the expectation that the model answers correctly, weighted by confidence (where $r(x)$ renormalizes to a [0, 1] range; see Appendix \ref{app:normalization}). Conversely, $1 - r (\E {g_t c_t})$ evaluates the model's calibration \textit{error} (MCE) on that question. High-confidence incorrect answers and low-confidence correct answers result in higher error, indicating poor calibration on question $q$. 

\subsection{Human-adjusted calibration error}\label{sec:metric-equation}
\metric{} incorporates human performance into MCE. This facilitates identification of specific cases where models are less calibrated than humans; rewards models more for being confidently correct before humans know the answer (simulating the competition setting); and penalizes them more for being confidently incorrect at that stage, placing greater weight on a higher-stakes error.

We thus weight the calibration at clue $t$ by $(1-h_t)$, the proportion of humans who have not yet answered correctly by clue $t$. A high score of $r (\E {(1-h_t) g_t c_t})$ indicates that the model is well-calibrated relative to human buzz performance.\footnote{A model is perfectly calibrated when it buzzes with full confidence ($c_t = 1$), is always correct ($g_t = 1$), and answers before humans buzz correctly ($h_t = 0$), resulting in $r (\E{(1 - h_t) g_t c_t}) = 1$.}
 Conversely, we estimate the expected probability for cases where the model does \textit{not} improve over humans, either due to incorrect answers or low confidence:
\begin{equation}
    \metric{}_q = 1 - r (\E{(1 - h_t) g_t c_t}).
\end{equation}
This adjustment evaluates the model's calibration error relative to human calibration performance on the same question.
We then define \( \metric{}_D \), the human-adjusted model calibration error for a benchmark \(D\), as the average of \( \metric{}_q \) across all questions.

%% file: 2025_acl_advcal/sections/60-calibration-evaluation.tex
\section{Model Calibration Evaluation}\label{sec:analysis}
\name{} helps to evaluate differences between human and model calibration (\S \ref{sec:analysis-buzz_compare}). We also validate the dataset's difficulty granularity and discuss calibration errors using our proposed metric (\S \ref{sec:analysis-metric}) and qualitative analysis (\S \ref{sec:qual_analysis}). 
%
%
\subsection{Comparing human and model calibration}\label{sec:analysis-buzz_compare}
\textbf{Buzz performance.}
\begin{figure}[!t] 
    \centering
    \includegraphics[width=\linewidth]{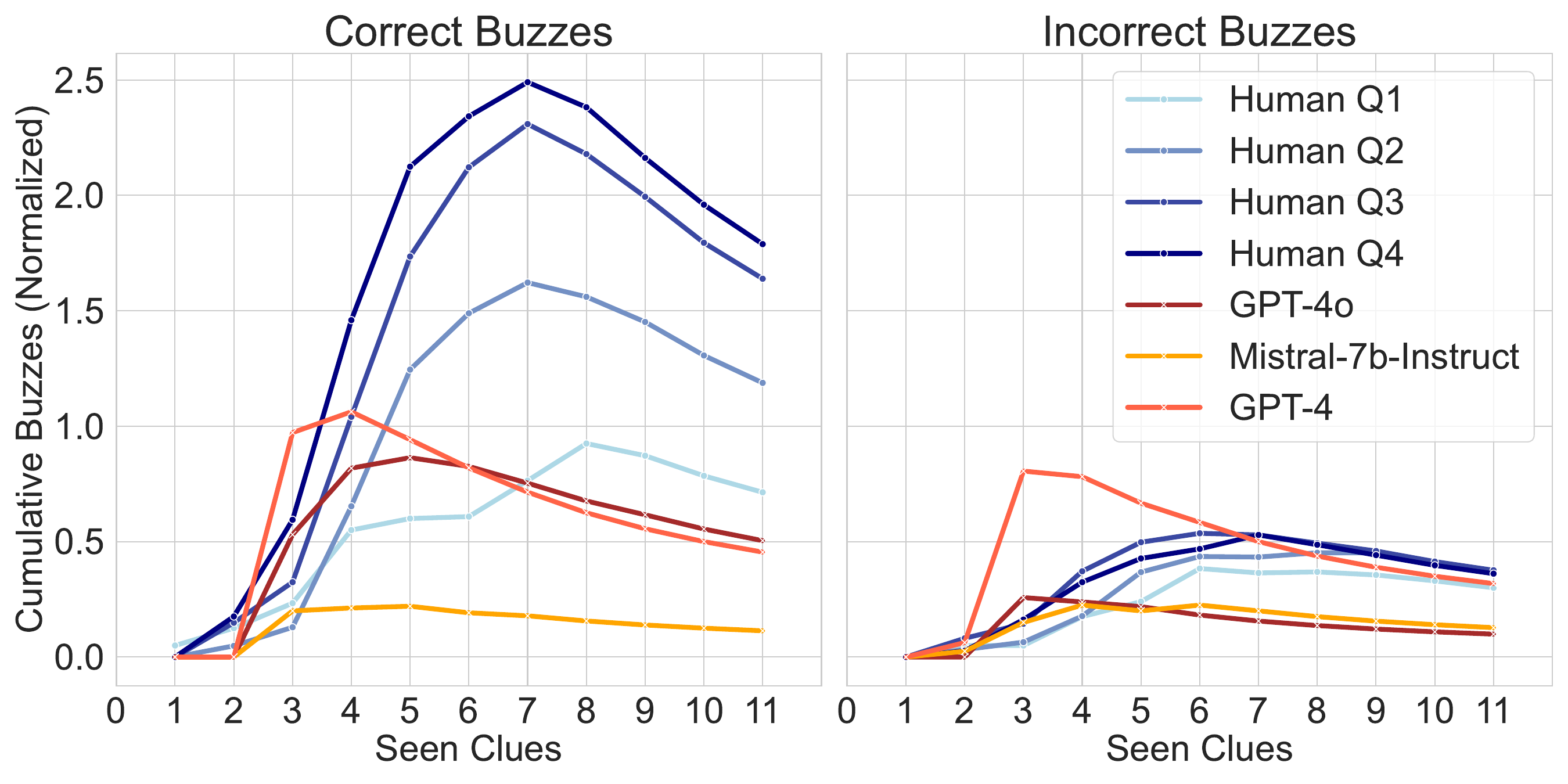}
    \caption{Each team's cumulative buzzes (normalized by the number of matches each team participated in). The top quartile of human teams (Q4) achieves the highest cumulative correct buzz rate, peaking over twice as high as the best model. Top human teams are thus more accurate and better-calibrated than models, even as the difficulty changes when more clues are revealed.}
    \label{fig:buzz_correctness_per_run}
\end{figure}
To compare human and model calibration, we first examine when and whether each team buzzes on the question, as well as the correctness of their answers. Figure~\ref{fig:buzz_correctness_per_run} gives each team's cumulative buzzes over the number of matches each team participated in. The 17 human teams are divided into quartiles, from Q1 (bottom) to Q4 (top), according to their total correct buzzes. Human teams, especially the top quartile, achieve the highest cumulative correct buzz rate (peaking in the middle of the questions), demonstrating their ability to confidently infer correct answers with fewer clues and indicating better accuracy and calibration than models. In contrast, GPT-4 exhibits a moderate cumulative correct buzz rate, which is only briefly higher than the top human teams and lower than 50\% of human teams for most of the question. Meanwhile, Mistral-7b-Instruct lags significantly behind all other teams, indicating poor calibration. In addition, GPT-4 and GPT-4o exhibit substantially higher incorrect buzz rates than human teams (right plot). All models, especially GPT-4, are overconfident early in the questions when little information is available: they are \textbf{especially miscalibrated relative to humans when the question is still hard}. Overall, the models tend to buzz incorrectly more often than humans and correctly less often, indicating \textbf{overconfidence in wrong answers and underconfidence in correct ones}.


\textbf{Difficulty granularity of each question.}\label{sec:analysis-adversarial}
\begin{figure}[!t] 
    \centering
    \includegraphics[width=\linewidth]{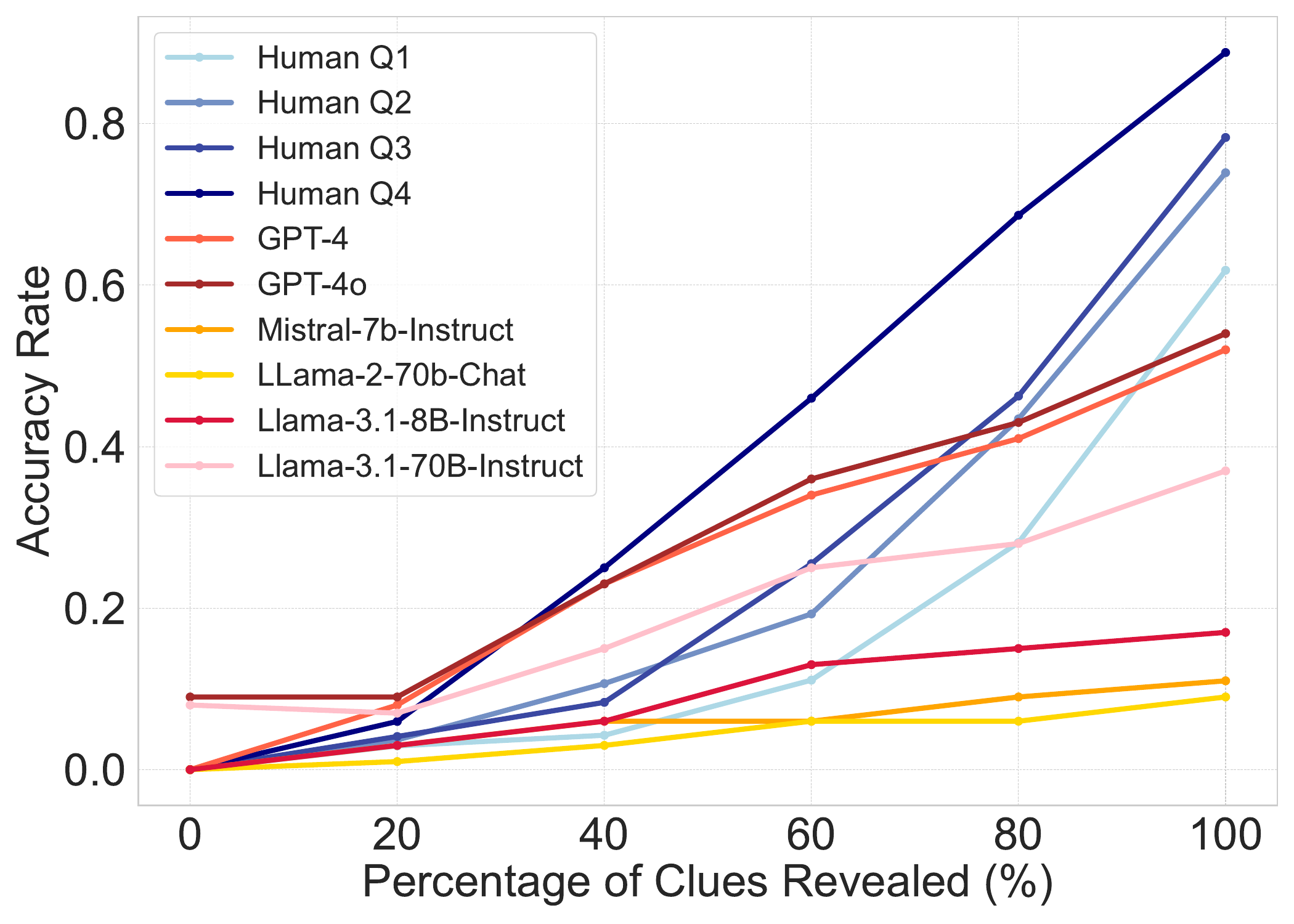}
    \caption{Comparison of human and model average accuracy rates as more clues are revealed (whether the team's guess is correct after seeing the first $n$ clues). As more clues are revealed, accuracy improves for both models and humans. Models often answer incorrectly until most clues are provided, and human accuracy increases more rapidly, validating that each instance becomes easier for both humans and models and that most humans can answer correctly by the end.}
    \label{fig:accuracy_rate}
\end{figure}
To evaluate model calibration over a range of difficulty levels for models, we asked the writers to write questions that are easier to answer as more clues are revealed (\S~\ref{sec:dataset}). To validate this design, 
we examine model and human correctness as the percent of clues revealed increases. For models, we consider the correctness of a model's guess for the first $n$ clues. The questions in \name{} are appropriately challenging and become easier for models and humans as more clues are revealed (Figure~\ref{fig:accuracy_rate}). Human team accuracy (blue) increases steadily, indicating that question difficulty indeed decreases as clues are revealed for human players. %
%
Moreover, even top models like GPT-4 and GPT-4o have under 50\% accuracy until at least 90\% of clues are provided, highlighting significant room for improvement on this benchmark.
%
To measure human accuracy per corresponding clue, we used offline human responses (\S \ref{sec:survey}; quartiles calculated per Appendix \ref{app:survey-quartiles}).

Notably, the bottom three quartiles of humans are less accurate than top models for most of the question (Figure \ref{fig:accuracy_rate}), yet still typically outperform models on maximizing correct buzzes relative to incorrect buzzes (Figure \ref{fig:buzz_correctness_per_run}). This trend suggests that \textbf{models' relatively high rate of incorrect buzzes and low rate of correct buzzes is due to miscalibration, not inaccuracy.} We investigate this distinction further in the next section.


\textbf{Conditional likelihood of correct answers.}
\label{sec:analysis-bayes}
\begin{figure}[!t] \centering\includegraphics[width=\linewidth]{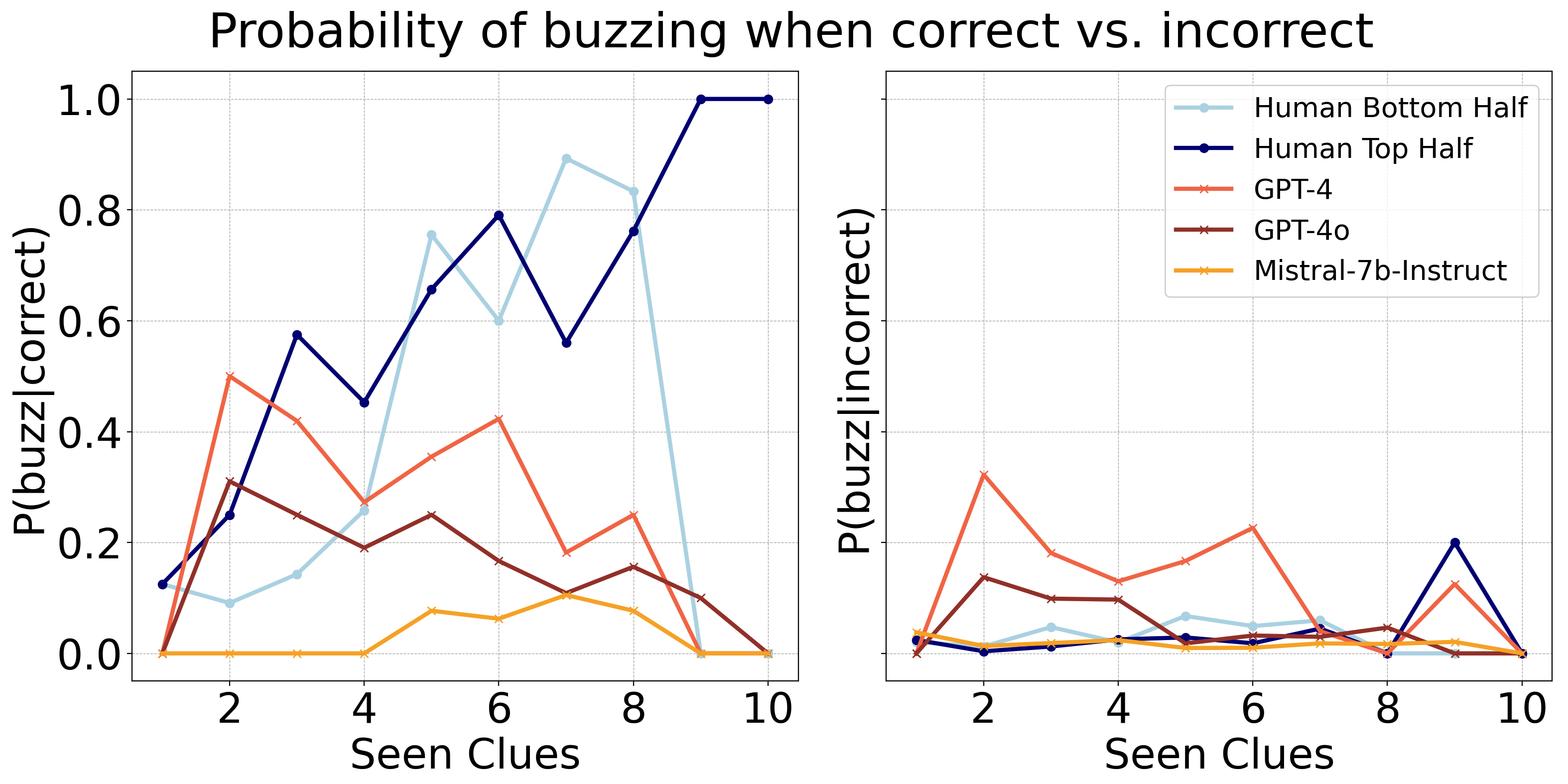}
    \caption{
    Humans are far more likely than models to buzz in when they are correct (left), and typically less likely to buzz in when they are incorrect (right), indicating that models remain miscalibrated relative to humans even when explicitly controlling for accuracy. (Due to the smaller sample size of human buzzpoints in the survey data, we use halves instead of quartiles here.)}
    \label{fig:bayes}
\end{figure}
While the tournament allows us to observe $P(g=1 \g b)$, the likelihood that a team's guess is correct ($g=1$) when they buzz ($b$), we also aimed to compare how often teams are confident enough to buzz when correct, $P(b \g g=1)$, and when incorrect, $P(b \g g=0)$. Using the offline human responses (\S \ref{sec:offline}), we estimate $P(b \g g=1)$ for each human team. For each player, we calculate $P_{player}(b \g g=1)$: the number of instances when a player buzzed in correctly with $n$ clues revealed, divided by the number of instances when a player's guess was correct with $n$ clues revealed. We then estimate $P(b \g g=1)$ for the top and bottom half of respondents (Appendix \ref{app:survey-quartiles}) as the average of $P_{player}(b \g g=1)$ across all surveyed players in that half. Finally, we compare this estimation with $P(g=1 \g b)$ for the tested models. We follow the same process to estimate $P(b \g g=0)$. \footnote{For all estimates, we consider only the guess correctness and buzz statistics up to the point when a player first buzzes, as later guesses do not count in real competitions.}
The results indicate that even \textbf{the strongest models are less confident than humans on correct answers and more confident on incorrect ones} (Figure \ref{fig:bayes}). For most questions, all humans are more than 50\% likely to buzz when correct, while models remain below 45\%, indicating lower confidence in correct answers. Among the models, GPT-4 was most likely to buzz incorrectly, reflecting its confidence in wrong answers.

As clues are revealed, humans become more likely to buzz when they know the correct answer, while models become less likely to buzz. This suggests that seeing more clues strengthens human confidence, but not model confidence.\footnote{A small fraction of human participants ($n$=1) has a sharp spike in incorrect buzzes near the end of a question.}






%% file: 2025_acl_advcal/sections/62-metric-analysis.tex
\subsection{\metric{} analysis}
\label{sec:analysis-metric}
\input{2025_acl_advcal/sections/metric-table}
We evaluate the calibration error of six LLMs on \name{} using existing metrics (ECE and Brier scores; Appendix~\ref{app:ece-brier}) and  \metric{}  (\S~\ref{sec:main-metric}). For each clue in a question, we collect  logit-based and verbalized confidences to compute metric scores. 

\metric{} correlates with ECE and Brier score results (Table~\ref{tab:calibration_table}); however, across both confidence elicitation methods, all models display greater error under human-adjusted \metric{} compared to MCE (\metric{} without human adjustment). \metric{} thus captures errors that existing methods overlook: cases where models underperform relative to humans by being confidently wrong or underconfident when correct. Thus, \textbf{\metric{} captures that models are especially ill-calibrated compared to humans}, and factoring in human performance reveals more room for improvement on LLM calibration. The gap between MCE and \metric{} widens for worse-performing models, suggesting that weaker models are even more miscalibrated relative to stronger models when factoring in human performance. Additionally, \metric{} reports higher errors than ECE and Brier scores across both confidence elicitation methods for most models, underscoring  calibration deficiencies that previous metrics underestimate.\footnote{All four metrics use a [0,1] scale; lower is better.} 

\subsection{Qualitative analysis and model errors}\label{sec:qual_analysis}
\textbf{Miscalibrated instances from \metric{}.}
All six models exhibit similar patterns for the questions on which they were most- and least-calibrated under \metric{} (examples in Appendix \ref{sec:poor-calib}). Models did best on questions that mention concrete proper nouns closely associated with the answer, even on obscure topics: for example, a question on \underline{Ireland} that gives the titles of Irish songs, a question on \underline{telomeres} that mentions the protein TRF2, and a question on \underline{Brooklyn} that mentions the neighborhood of Midwood. Models tend to be least-calibrated on questions with multiple plausible answers (e.g., one on \underline{fish} as a Buddhist symbol, since other animals also have symbolic meanings in Buddhism). Models also struggle on questions that use descriptions instead of titles (e.g. a question that describes music by \underline{Maurice Ravel}, and one that describes Jewish \underline{birth} ceremonies).

\textbf{Qualitative feedback.} We survey the human players for feedback on model abilities. Differences in model calibration are visible to the players: several find GPT-4 ``too aggressive,'' while Mistral seems much weaker, often buzzing late in the question. One player notes that the models ``obviously knew a lot, but were quite bad at gauging how well they knew something to [buzz].'' Others note that models tend to buzz on ``more concrete clues'' and struggle with multi-step reasoning. For example, a question on \underline{Alice Walker} mentions her trip to Eatonville to write about local author Zora Neale Hurston. Players note that GPT-4 incorrectly guesses ``Zora Neale Hurston,'' while human players correctly say ``Alice Walker.''

Players also note that when models were incorrect, they give more ``unreasonable'' answers than humans do. For example, models incorrectly answer a question on the treatise \underline{Philosophical Investigations} with ``Fermat's Little Theorem'' and ``\textit{The Lion, the Witch and the Wardrobe}.'' Guessing an equation and a children’s book with high confidence for a work of philosophy suggests serious miscalibration, since either option should be completely outside the realm of possibility; no human players gave answers so distant from the correct one. Other model errors not observed among human players include buzzing before any substantive clues are revealed, answering with a song title for a question asking for a surname, and hallucinating inexistent schools of philosophy.

Model and human strengths between question topics differ greatly. We examine models' and humans' ratios of correct to incorrect buzzes per category. Human players are best at literature, but this is the weakest or second-weakest category for all models. All models did relatively well on science. GPT-4o is much stronger at social science, arts, and science than other categories, and slightly outperforms humans for every category; GPT-4 was worse than the humans for all categories.

All human participants in our competitions were experienced players, but we find that calibration performance varies greatly even among these experts: stronger humans substantially outperform top models, but not all humans do. A general takeaway for future model-human comparisons on tasks involving calibration is that variance in human skill can greatly affect the outcome of a comparison.

A side benefit of conducting live human-model competitions was a significant degree of community involvement from trivia enthusiasts who were not researchers. In-person data collection, though more involved than crowdsourced data, offers other benefits: we found that participants were attentive and enthusiastic; moreover, in-person data collection (especially ``gamified’’ approaches) raises awareness of and interest in AI.

%% file: 2025_acl_advcal/sections/metric-table.tex
\begin{table}[t]
\footnotesize
\centering
\setlength{\tabcolsep}{5pt}
\renewcommand{\arraystretch}{1.3} 
\resizebox{\linewidth}{!}{
\begin{tabular}{lcccc}
\toprule
\multicolumn{5}{c}{\textit{Verbalized-based Confidence}} \\
\midrule
\textbf{Model} & \textbf{Brier Score} & \textbf{ECE} & \textbf{MCE} & \textbf{\metric{}} \\
\midrule
GPT-4 & \cellcolor{Violet!20}0.274 \rankbox{2} & \cellcolor{Violet!20}0.259 \rankbox{2} & \cellcolor{RoyalPurple!20}0.584 \rankbox{1} & \cellcolor{RoyalPurple!20}0.588 \rankbox{1} \\
GPT-4o & \cellcolor{RoyalPurple!20}0.266 \rankbox{1} & \cellcolor{RoyalPurple!20}0.224 \rankbox{1} & \cellcolor{Violet!20}0.601 \rankbox{2} & \cellcolor{Violet!20}0.604 \rankbox{2} \\
Llama-3.1-70B-Instruct & \cellcolor{Purple!20}0.373 \rankbox{3} & \cellcolor{Purple!20}0.392 \rankbox{3} & \cellcolor{Purple!20}0.685 \rankbox{3} & \cellcolor{Purple!20}0.719 \rankbox{3} \\
LLama-2-70b-Chat & \cellcolor{WildStrawberry!20}0.490 \rankbox{4} & \cellcolor{WildStrawberry!20}0.570 \rankbox{4} & \cellcolor{WildStrawberry!20}0.739 \rankbox{4} & \cellcolor{WildStrawberry!20}0.803 \rankbox{4} \\
Llama-3.1-8B-Instruct & \cellcolor{RedOrange!20}0.623 \rankbox{5} & \cellcolor{RedOrange!20}0.693 \rankbox{5} & \cellcolor{RedOrange!20}0.774 \rankbox{5} & \cellcolor{RedOrange!20}0.843 \rankbox{5} \\
Mistral-7b-Instruct & \cellcolor{YellowOrange!20}0.716 \rankbox{6} & \cellcolor{YellowOrange!20}0.784 \rankbox{6} & \cellcolor{YellowOrange!20}0.790 \rankbox{6} & \cellcolor{YellowOrange!20}0.881 \rankbox{6} \\
\midrule
\multicolumn{5}{c}{\textit{Logit-based Confidence}} \\
\midrule
\textbf{Model} & \textbf{Brier Score} & \textbf{ECE} & \textbf{MCE} & \textbf{\metric{}} \\
\midrule
GPT-4o & \cellcolor{Purple!20}0.341 \rankbox{3} & \cellcolor{Violet!20}0.353 \rankbox{2} & \cellcolor{RoyalPurple!20}0.654 \rankbox{2} & \cellcolor{RoyalPurple!20}0.661 \rankbox{1} \\
Llama-3.1-70B-Instruct & \cellcolor{Violet!20}0.323 \rankbox{2} & \cellcolor{RoyalPurple!20}0.339 \rankbox{1} & \cellcolor{Violet!20}0.651 \rankbox{1} & \cellcolor{Violet!20}0.679 \rankbox{2} \\
GPT-4 & \cellcolor{WildStrawberry!20}0.380 \rankbox{4} & \cellcolor{Purple!20}0.388 \rankbox{3} & \cellcolor{Purple!20}0.672 \rankbox{3} & \cellcolor{Purple!20}0.684 \rankbox{3} \\
Llama-3.1-8B-Instruct & \cellcolor{RoyalPurple!20}0.302 \rankbox{1} & \cellcolor{WildStrawberry!20}0.397 \rankbox{4} & \cellcolor{WildStrawberry!20}0.675 \rankbox{4} & \cellcolor{WildStrawberry!20}0.718 \rankbox{4} \\
Mistral-7b-Instruct & \cellcolor{RedOrange!20}0.553 \rankbox{5} & \cellcolor{RedOrange!20}0.677 \rankbox{5} & \cellcolor{RedOrange!20}0.766 \rankbox{5} & \cellcolor{RedOrange!20}0.846 \rankbox{5} \\
Llama-2-70b-Chat & \cellcolor{YellowOrange!20}0.774 \rankbox{6} & \cellcolor{YellowOrange!20}0.829 \rankbox{6} & \cellcolor{YellowOrange!20}0.825 \rankbox{6} & \cellcolor{YellowOrange!20}0.921 \rankbox{6} \\
\bottomrule
\end{tabular}
}
\caption{Across both confidence elicitation methods, all models display greater error under human-adjusted \metric{} compared to MCE (\metric{} without human adjustment). \metric{} thus captures errors that existing methods overlook: cases where models underperform relative to humans by being confidently wrong or underconfident when correct.}
\label{tab:calibration_table}
\end{table}

%% file: 2025_acl_advcal/sections/70-conclusion.tex
\section{Conclusion}\label{sec:conclusion}

For users to trust LLMs, they need assurance that
these models will not confidently produce wrong answers. To address this, \name{} offers a benchmark for fine-grained calibration evaluation,
grounded in human calibration. Our analyses on
\name{} reveal that models are often miscalibrated relative to
well-informed humans. Specifically, model calibration errors came from difficulty with abstract descriptions, far-fetched incorrect guesses, and confidently incorrect answers given few clues.
%
Our new metric, \metric{}, combined with \name{}, evaluates the performance of six LLMs, revealing significant room for improvement.

%
%
\name{} provides a blueprint for developing human-focused improvements to calibration: improving verbalized confidences that measurably help human decision-making, personalizing abstention based on individual human skill, and measuring these interactions via human-model teaming. 

%% file: 2025_acl_advcal/sections/80-limitation.tex
\section*{Limitations}
Since \name{} focuses on a question-answering task, its applicability to broader NLP domains remains unexplored. Future work should explore calibration in other open-ended generation settings to assess generalizability. Furthermore, while our proposed metric, \metric{}, serves as a baseline, it is not exhaustive of all forms of uncertainty and miscalibration. For instance, enhancing this metric for human-AI collaboration could help users determine when to rely on models and when to defer to human judgment.

\section*{Acknowledgments}

Many thanks to the question editors who helped to create the dataset: Ankit Aggarwal, Jaimie Carlson, Bradley Kirksey, Linus Luu, Shahar Schwartz, Noah Sheidlower, and Jonathan Tran. Special thanks to the writers who assisted as well. This research was only possible due to the generous participation of the players who competed in the tournaments (special thanks to Joy An, Andrew Gao, Kevin Yu, Alex Stonemanxa, Priyanka Raghavan, and Magdalena Lederbauer) and the moderators who read the questions. We thank Lakshya Agrawal, Nishant Balepur,  Nicholas Tomlin, Swabha Swayamdipta, and Irene Ying for providing feedback on earlier drafts.

This research was partly supported by an NSF GRFP grant.

%% file: 2025_acl_advcal/sections/appendix.tex
\section{Ethical Considerations} \label{Ethics}
We address ethical considerations for dataset papers, given that our work contains a new dataset, \name{}, and collects human responses in our user study. We reply to the relevant questions posed in the {\texttt{\abr{acl} 2022 Ethics \abr{faq}}}\footnote{\url{https://www.acm.org/code-of-ethics}}. 

When collecting human responses and questions, our study was pre-monitored by an official \abr{irb} review board to protect the participants' privacy rights. Buzzpoints in \name{} are anonymous; the feedback survey asked for respondents’ names for purposes of compensation, but only aggregate data and anonymous quotes are used in our study. 
Before distributing the survey, we collected consent forms for the workers to agree that their answers would be used for academic purposes. 
The trivia players were awarded a total $\$900$ \abr{USD} worth of online gift cards after the competitions. The prizes were \$150, \$100, \$50 for the first three places at each site where the tournament was held. 
The trivia writers were paid \$5 per question and editors paid \$2.50 per question edited based on the estimated completion time, and which was calculated to reach over $\$10$ \abr{usd} an hour (a rate higher than the \abr{us} national minimum wage of $7.50$ \abr{usd}
).


\section{Dataset Details}
\label{app:dataset-design}

\subsection{Distribution of questions in dataset}
Questions in \name{} are distributed as follows:
\begin{itemize}
    \item 20\% literature: 5\% American literature; 5\% British literature; 5\% European literature; 5\% world and other literature
    \item 20\% history: American history; 5\% world history; 5\% European history; 5\% other history
    \item 20\% science: 5\% Biology; 5\% Chemistry; 5\% Physics; 5\% computer science, math, and other science
    \item 15\% arts: 5\% painting/sculpture; 5\% classical music; 5\% other arts
    \item 15\% social sciences: 5\% religion; 5\% philosophy; 5\%
    \item 5\% geography and current events
    \item 5\% myth, pop culture, and other
\end{itemize}

This distribution is based on the standard quizbowl distribution at \texttt{acf-quizbowl.com/distribution}.

\subsection{Sample dataset questions}
\qbquestion{The protagonist of the story “Storm in a Teacup” obsessively reads a book with this number in its title. This number partly titles a work that opens by noting that people are born naturally good, used to teach children. One hundred times this number titles a poetry anthology whose length was inspired by the Classic of Poetry. A man visits a hut this number of times to recruit the Sleeping Dragon, and this many characters pledge to die on the same day while swearing the Peach Garden Oath. For 10 points, a Luo Guanzhong novel is titled for how many kingdoms?}
{\underline{three} [or \underline{san}; accept \underline{Three} Character Classic or \underline{Three} Hundred Tang Poems or Romance of the \underline{Three} Kingdoms; accept \underline{Sān}zì Jīng or Tángshī \underline{sān}bǎi shǒu or \underline{Sān}guó Yǎnyì]}
\vspace{10pt}

\qbquestion{During this decade, some members of the Circle of Seven participated in a group called the Golden Square and led a coup to remove the Iraqi government from power. The Pahlavi ruler and father of Mohammed Reza Shah was forced out of power in this decade, during which a supply route was developed to run through port cities to Tehran. The Syria-Lebanon campaign of this decade was led by Archibald Wavell, who later lost in the Western Desert, where the Battle of El Alamein took place. For 10 points, General Erwin Rommel fought in what decade when most of World War Two occurred?}
{\underline{1940s} [prompt on 40's]}
\vspace{10pt}
\qbquestion{One form of this adjective describes a distributed object whose state is equivalent to a strictly serializable database. Weight decay tends to pull a sigmoid activation function towards this adjective's namesake "regime". If all activation functions of a deep neural network have this form, the output is provably this type of mapping of the inputs. ReLU is a modified piecewise form of this type of function, which is the simplest separator in two-dimensional binary classification. The most common shape of least-squares regression is described by, for 10 points, what adjective that describes functions of the form “y=mx+b”?}
{\underline{linear} [accept \underline{linearizable} or word forms; prompt on planar or hyperplanar or word forms]}
\vspace{10pt}
\qbquestion{The NSF’s IDP sets standards for tools used to extract this substance, which include Foro 3000 and DISC. Coral, certain protists, and this substance are the primary sources of delta-O-18 information. Very thin layers of this substance that look like oil spills are nicknamed for grease. Due to diffusion, clathrates and trapped air in this substance help to track atmospheric gas concentrations, and may be retrieved from masses that undergo calving, resulting in growlers or bergy bits. For 10 points, name this material, which may be extracted in core samples from namesake sheets or from glaciers.}
{\underline{ice} [or \underline{frozen water}; or \underline{solid water}; accept \underline{ice} sheets; accept \underline{icebergs}; prompt on snow or firn; prompt on glaciers; prompt on water}
\vspace{10pt}
\qbquestion{Sephardic Jews have recently moved into this region near the original Vitagraph studios in Midwood. The world headquarters of the Chabad (“huh-BAHD”) movement is located in this region within a neighborhood where the Rebbe struck and killed Gavin Cato in 1991. Many Jews of the Bobov sect live in this region's neighborhood of Borough Park. This region, where Jews clashed with Black residents in the Crown Heights Riots, is where many Hasidic Jews live in the increasingly trendy neighborhood of Williamsburg. For 10 points, name this New York City borough whose Jewish residents often live near Coney Island.}
{\underline{Brooklyn} [or \underline{Kings County}; prompt on New York City]}
\vspace{10pt}
\qbquestion{At the 2023 Women’s World Cup, the Spanish and Dutch teams sparked controversy by seemingly performing this dance during training. Lyrics commonly sung to this dance describe putting one foot in front of the other “until the Sun shines on me.” The currently youngest lawmaker in one country led this dance while tearing papers in half. Participants in a November 2024 march performed this dance in protest over a bill introduced by David Seymour that would affect the Treaty of Waitangi. For 10 points, recent protests in New Zealand have made use of what Māori ceremonial dance?}
{\underline{haka}}

\subsection{Sample questions for calibration analysis}
\label{sec:poor-calib}

Sample question on which models are poorly calibrated:

\qbquestion{One thinker's argument that this claim is a "hyperbolic point which ought to be silent" was subject to a response titled "My Body, This Paper, This Fire." Jacques Derrida first coined the word "diff\'erance" in a book responding to Michel Foucault's Madness and Civilization and partially titled for this statement. In The Search for Natural Light, a premise involving doubt was added to this statement. This claim, which Pierre Gassendi criticized for being circular, was presented as an example of a "clear and distinct" idea. For 10 points, name this first principle coined in Ren\'e Descartes' Discourse on the Method.}
{\underline{"I think, therefore I am"} [or \underline{"cogito, ergo sum"}]} 

Sample question on which models are well-calibrated:
\label{sec:well-calib}

\qbquestion{Sephardic Jews have recently moved into this region near the original Vitagraph studios in Midwood. The world headquarters of the Chabad movement is located in this region within a neighborhood where the Rebbe struck and killed Gavin Cato in 1991. Many Jews of the Bobov sect live in this region's neighborhood of Borough Park. This region, where Jews clashed with Black residents in the Crown Heights Riots, is where many Hasidic Jews live in the increasingly trendy neighborhood of Williamsburg. For 10 points, name this New York City borough whose Jewish residents often live near Coney Island.}
{\underline{Brooklyn} [or \underline{Kings County}; prompt on New York City]}



\section{Dataset Comparison for Adversarialness}
\label{app:trickme-comparison}
We compare \name{} with the TrickMe dataset~\cite{wallace2019trick} on
(1) question length and
(2) question difficulty evaluation.
Overall, \name{} questions are longer: \name{} has an average of 5.09 sentences per question, while TrickMe has an average of 4.74 sentences.
We compare the difficulty based on the accuracy rate as the clues are revealed in Table~\ref{tab:trickme-comparison}. \name{} exhibits higher adversarialness throughout the incremental questions.

\begin{table}
\resizebox{\linewidth}{!}{
\centering
\begin{tabular}{@{}llccccc@{}}
\toprule
 &  & \textbf{20\%} & \textbf{40\%} & \textbf{60\%} & \textbf{80\%} & \textbf{100\%}\\ \midrule
\multirow{2}{*}{\textbf{TrickMe}} & \textsc{GPT-4} & 31.07\% & 65.97\% & 82.20\% & 89.96\% & 93.68\% \\
 & \textsc{GPT-4o} & 28.77\% & 61.56\% & 77.90\% & 86.64\% & 91.11\% \\
 \midrule
\multirow{2}{*}{\textbf{\name{}}} & \textsc{GPT-4} & 7.49\% & 23.62\% & 30.60\% & 40.40\% & 52.19\% \\
 & \textsc{GPT-4o} & 8.59\% & 22.65\% & 32.05\% & 42.86\% & 53.78\% \\
 \bottomrule
\end{tabular}
}
\caption{\name{} and TrickMe dataset accuracy rate comparison. Similar as Figure~\ref{fig:accuracy_rate}, we compare the accuracy when the percentage of clues revealed (\%) are 20\%, 40\%, 60\%, 80\%, 100\%.
\name{} results show much lower accuracy compared with TrickMe, indicating that our dataset is more adversarial incrementally.
}
\label{tab:trickme-comparison}
\end{table}

\section{Interface}
\subsection{Guesser details}\label{sec:guesser}
Before deciding when to buzz, models need to generate answers as clues are revealed. We call this process ``guessing'' and the model is used as a \textit{guesser}.
The full prompt of the guesser is in Appendix~\ref{app:buzzpoint-prompt}, including the general instructions and retrieved examples.
We train a TF-IDF model as the retriever with past quizbowl questions following~\citet{rodriguez2019quizbowl}.\footnote{\url{https://scikit-learn.org/stable/modules/generated/sklearn.feature_extraction.text.TfidfVectorizer.html}}
The main goal is to reduce hallucinations and guide the model to learn the granular clue and guess format.


\subsection{Training buzzer for authoring interface}\label{logistic}
For the writing interface, the logistic regression model was trained with two kinds of features: GPT-3.5's logit based confidence, and pre-designed features derived from the question, answer, and metadata. Features include text-based metrics (e.g., TF-IDF scores, overlaps between Llama predictions and TF-IDF guesses), probabilistic outputs (Llama log and prompt probabilities), and contextual indicators (sentence index, length, and presence of phrases like ``10 points''). This model was trained on \citet{rodriguez2019quizbowl}, also pyramidal questions, using Llama-13b predictions~\citep{touvron2023llama}.

A primary
distinction from \citet{wallace2019trick} is that their interface only showed
the final correct guess.

\subsection{Question editors and writers}
\label{app:writer-qualifications}
Writers and editors were recruited via a public quizbowl forum and were located in the US and UK. All editors underwent IRB training and had written for at least three previous tournaments. Writers were paid \$5 per question and editors paid \$1 per question edited. A head editor with 5 years of experience writing and editing quizbowl questions, including as head editor of two previous tournaments, supervised the writers and provided an additional quality check.
The consent form is in Table~\ref{tab:consent-for-writers}.

\begin{table*}[t]
\centering
\resizebox{\textwidth}{!}{
\begin{tabular}{@{}c@{}}
\toprule
\begin{tabular}[c]{@{}l@{}}
\textbf{Privacy Policy}\\
\midrule
\textbf{Introduction}\\
Welcome to 'Stump the Computer,' hosted by the research group at \texttt{xxx}. Your participation in our research is voluntary and deeply valued. \\ This Privacy Policy outlines how we handle the information you provide during your interaction with our research project.\\
\textbf{Purpose of the Study}\\
This research, conducted by \texttt{xxx}, aims to collect human-generated data for a fact verification system. \\ Participants, like you, who enjoy trivia knowledge and trivia-related games, are invited to help us understand how humans and computers handle challenging questions.\\
\textbf{Information Collection and Use}\\
As you participate in this project, you will interact with an online interface to submit questions designed to challenge both human and computer intelligence. \\You may use a Google Doc, automatically generated in your Google Account, to draft these questions. \\We collect this data to enhance our research and improve the interaction models between humans and computers.\\
\textbf{Data Confidentiality}\\
We take your privacy seriously. All data collected during this study will be stored on a password-protected web server with encrypted storage. \\The server is in a secure access data center, managed professionally. Access to the data is strictly limited to the principal investigators of this study. \\After the study concludes, any personally identifiable information will be anonymized or destroyed to ensure your privacy.\\
\textbf{Potential Risks}\\
We acknowledge the risk of breach of confidentiality in any online activity. \\We have implemented robust security measures to mitigate such risks and protect your data.\\
\textbf{Benefits and Compensation}\\
While there is no direct personal benefit from participating, your contributions are invaluable in advancing research on human-computer interactions. \\Compensation for participating includes \$5 per question written and \$2.50 for each question edited.\\
\textbf{Your Rights}\\
Participation is entirely voluntary. You may withdraw at any time without penalty. Should you have any questions or need to report a concern, please contact:\\
\texttt{xxx}\\
\textbf{Changes to This Policy}\\
We may update this policy periodically to reflect changes in our practices. Continued participation after such changes constitutes your acceptance of the new terms.\\
\textbf{Consent}\\
By signing up and participating, you affirm that you are at least 18 years old, have reviewed this policy, and consent to engage in this study. \\Your rights and privacy are paramount, and we are committed to protecting them.\\ \\
Thank you for participating in our task and contributing to the advancement of our research.\end{tabular} \\ \bottomrule
\end{tabular}
}
\caption{Consent form for question writers.}
\label{tab:consent-for-writers}
\end{table*}

\subsection{Model confidence elicitation}\label{sec:confidence_elicitation}
We experiment two approaches to get the model confidence score for a generated guess.

\paragraph{Token-logit based confidence} Log probability of the generation is a common
method to estimate the model confidence~\cite{nguyen-oconnor-2015-posterior}.
To get the confidence score in our setup, we retrieve the logit for each generated token, and take the average of the exponentials of these logit values~\citep{huang2023look}.

\paragraph{Verbalized confidence} Recent study shows verbalized probabilities
can be better calibrated than log probabilities~\cite{tian-etal-2023-just, xiongcan}, which motivate us to include the verbalized confidence in our experiments.
We follow the previous prompt in Appendix~\ref{app:buzzpoint-prompt} and add the instructions from~\citet{tian-etal-2023-just} to return the confidence:
    \begin{quote}
        Given the following information, provide the title of the Wikipedia page that would best answer the last question fragment. If you are not sure, just give your best guess. If you don't know, answer None. The answer should be as short as possible. \textbf{While you give the guess, please also provide the probability that it is correct (0.0 to 1.0).} \\
        \textbf{Give ONLY the guess and probability, no other words or explanation. For example:}

        \textbf{The answer is: <most likely guess, as short as possible; not a complete sentence, just the guess!>} \\
        \textbf{Probability: <the probability between 0.0 and 1.0 that your guess is correct, without any extra commentary whatsoever; just the probability!>}

        Question: \{retrieved examples\} \\
        The answer is: \{retrieved examples\}
        
        Question: \{each clue\}
    \end{quote}

To ensure accurate extraction of probability scores from model outputs, we initially define the desired format based on the prompt. We then proceed to identify and print any cases where confidence scores are not successfully extracted. By observing these cases, we can discern patterns and refine our post-processing rules. This iterative approach allows us to capture as many corner cases as possible, enhancing the robustness of our data extraction process.



\section{Model Buzzpoint Generation}
\subsection{Guesser details}\label{app:buzzpoint-prompt}
To retrieve a model's top guess after $n$ clues have been revealed, we prompt the model with the first $n$ clues from the question. To retrieve the best guess, we employ a "retrieval and guess" approach to enhance QA performance, using the following prompt:
    \begin{quote}
        Given the following information, provide the title of the Wikipedia page that best answers the last question fragment. If unsure, provide your best guess. The answer should be concise.

        Question: \{retrieved examples\} \\
        The answer is: \{retrieved examples\}
        
        Question: \{each clue\} \\
        The answer is:
    \end{quote}

\subsection{Process for calculating model buzzpoints}\label{app:buzzpoint-algo}
The process for calculating model buzzpoints followed the steps below.
\begin{algorithm}
\small{
\begin{algorithmic}
\caption{Find model \buzzpoints{} for a question}
\label{alg:feature}

\Require $N$, the number of clues in the question; and $t$, the buzz threshold.
\State Let $n = 0$ 
\While{$n < N$}
\State {Prompt the model to answer the question, given the first $n$ clues (Appendix \ref{app:buzzpoint-prompt}).}

\State Compute the model's confidence $c$ in its top guess by summing the log probabilities of the tokens comprising the guess.

\If{$c$ > $t$}
\State Buzz in
\State{\textbf{break}}
\EndIf
\State $n$ += 1
\EndWhile
\end{algorithmic}
}
\end{algorithm}

\subsection{Assigning buzzer threshold}\label{app:buzzpoint-threshold}
We used question data from the 2023 Expo quizbowl competition, where expert players competed against ChatGPT as a testbed to assign buzzer threshold. The dataset includes questions covering various topics, with recorded buzz and guess correctness.

We incorporate the explanation from \citet{rodriguez2019quizbowl} to explain how the threshold was set for our buzzer.
To estimate the probability $\pi(t)$ that a player has answered correctly by position $t$, the following formula is used:

\begin{equation}
    \pi(t) = 1 - \frac{N_t}{N},
\end{equation}

where $N$ is the total number of player-question records, and $N_t$ is the number of instances where a player has answered correctly by position $t$. 
This equation indicates how likely it is that a player has given the correct answer by a certain point in the question.
To make this probability easier to use in practice, it is approximated with a polynomial function:
\begin{equation}
    \pi(t) = 0.0775t - 1.278t^2 + 0.588t^3.
\end{equation}

This polynomial provides a smooth estimate of how human accuracy changes as the question progresses, 
allowing for a data-driven approach to determining optimal buzz thresholds.
Since we also want to get the optimal buzzpoints, we adopt a threshold from this study to determine the model buzzpoints. The confidence thresholds by model family are: -0.03 for GPT models, and  -0.05 for Mistral models.

\section{Tournament and Survey}
\label{app:tournament}
\subsection{Recruiting human players}\label{players}
Human teams were recruited by posting a call for players on social media and public forums for quizbowl players. To incentivize teams to play as well as possible, the top three teams in each tournament were awarded a prize. The prizes were \$150, \$100, \$50 for the first three places at each site where the tournament was held. The tournaments were all held in the US and the players were also located in the U.S. 
The consent form is in Table~\ref{tab:consent-for-players}.

\begin{table*}[t]
\centering
\resizebox{\textwidth}{!}{
\begin{tabular}{@{}c@{}}
\toprule
\begin{tabular}[c]{@{}l@{}}
Welcome to a quick Trivia Quiz! You'll tackle 37 questions and decide if you're confident enough to buzz. \\After each clue, please write down your best guess even if you're not sure of the answer. \\ \\

To track your progress, refer to the progress bar at the top of the page (Please disregard the question numbers—they are randomized for fairness). \\ \\

Please enter your email address here so we can ensure prizes are sent to the right recipients at the end of the quiz. \\We're giving away 4 raffle prizes (each a \$25 gift card) to participants who complete the quiz. \\Additionally, the top scorer with the highest accuracy and fastest buzz will receive a \$5 bonus prize. \\Rewards are determined by a combination of your accuracy and buzzing speed. \\Correct answers receive 0 to 20 points depending on how early you buzz. Incorrect answers receive -5 points. \\Good luck and have fun! \\ \\

Here is the short summary of what this survey will be used for: \\

\textbf{Project Title}\\
A Leaderboard and Competition for Human-computer Adversarial Question Answering \\

\textbf{Purpose of the Study}\\
This research is being conducted by \texttt{xxx} at \texttt{xxx}. \\We are inviting you to participate in this research project because you are interested in trivia knowledge and enjoy trivia-related games. \\The purpose of this research project is to collect human-generated data and responses for building a deployable QA system. \\

\textbf{Procedures} \\
We would like to use two adversarial datasets to test our approach of determining how adversarial the questions are. \\You will provide your best guess and write down your answer in a short form.\\

\textbf{Potential Risks and Discomforts}\\
There is a risk of breach of confidentiality and that efforts to mitigate this risk are described in the Confidentiality section below. \\

\textbf{Potential Benefits} \\
While this research is not designed to benefit you personally, we hope that in the future, the researchers might benefit from human computer interaction studies \\with investigation of \\question answering behavior and AI development during this study. \\

\textbf{Confidentiality}\\
Only the investigators (\texttt{xxx}) of this study will have access to the study data. \\Data will be stored on a password-protected web server with encrypted storage. \\The server is professionally managed in a secure access data center. \\After the study ends, only user names associated with e-mail addresses will be retained and the associated e-mail addresses will be deleted. \\

\textbf{Right to Withdraw and Questions} \\
Your participation in this research is completely voluntary. You may choose not to take part at all. \\If you decide to participate in this research, you may stop participating at any time. \\If you decide not to participate in this study or if you stop participating at any time, you will not be penalized or lose any benefits to which you otherwise qualify. \\If you decide to stop taking part in the study, if you have questions, concerns, or complaints, or if you need to report an injury related to the research, please contact \\the investigator: \\
\texttt{xxx} \\ \\

If you have any questions, please use this address: \texttt{xxx}
\end{tabular} \\ \bottomrule
\end{tabular}
}
\caption{Consent form for players.}
\label{tab:consent-for-players}
\end{table*}

\subsection{Player expertise}\label{app:player_specifics}
Respondents had an average of 5.5 years of previous experience playing quizbowl. 22\% of players had studied or were currently studying in the physical sciences or engineering; 31\% studied computer science or math; 17\% studied the humanities; 13\% studied a combination of fields; and 17\% were undecided. Since quizbowl players typically specialize in certain categories and learn more about those areas, we also asked them for their areas of specialization. 39.13\% of respondents listed the sciences as an area of specialization; 21.74\% listed history; 39.13\% listed the social sciences; 52.17\% listed literature; 39.13\%	listed fine arts; and 21.74\% listed geography or current events.

\subsection{Ranking human performance from survey}
\label{app:survey-quartiles}
Because the survey measures individual accuracy, without a competition setting, we rank humans using the following metric: participants earn $(20-20c)$ points per question with a correct buzz and lose 5 points per question with an incorrect buzz, where $c$ is the proportion of clues seen at the buzzpoint. Human quartiles and top/bottom half categorization based on the offline buzzpoints (for Figures \ref{fig:accuracy_rate} and \ref{fig:bayes}) are taken from rankings under this metric.

\section{ECE and Brier Score Details}
\label{app:ece-brier}
Expected Calibration Error (ECE)~\citep{naeini2015obtaining, guo2017oncalibration} and Brier scores \cite{brier1950verification} are widely used metrics for assessing model calibration.
Below, we define these metrics using the notations introduced in Section~\ref{sec:baseline-metric}:
\( g_t \) represents the answer correctness at clue \(t\) ($1$ if correct, $0$ otherwise, which is slightly different from \metric{} for simplicity);
\( c_t \) represents the corresponding model's confidence in its guess.

\paragraph{ECE} It measures the weighted average over the absolute difference between accuracy and confidence. To compute this, we first split confidence values into \( M = 10 \) bins equally.
\( B_m \) represents the confidence set of the \(m^{\text{th}}\) bin.
\( N \) is the total number of clues across the dataset.
\[ \text{ECE} = \frac{1}{N} \sum_{m=1}^M \bigg| \sum\limits_{t \in B_m} g_t - \sum\limits_{t \in B_m} c_t \bigg| \]

\paragraph{Brier Score} It measures the mean squared difference between the predicted probability and the actual binary outcome, measuring how well the predicted confidence aligns with the true correctness of the answer. A lower Brier Score indicates better calibration, as it reflects more accurate and well-calibrated probability estimates.
\[ \text{Brier Score} = \frac{1}{N} \sum_{t=1}^N (c_t - g_t)^2 \]

\section{Another Potential Human-grounded Calibration Metric using \name{}}
We use the same notation as \S~\ref{sec:metric-equation}. To quantify model calibration, we define the \textit{buzz confidence}, \( b_t \), representing the likelihood that the model will be confident at time step \( t \). This probability depends on the model's confidence \( c_t \) at \( t \) and the cumulative non-buzz probabilities from earlier steps, expressed as \( b_t = c_t \prod_{i=0}^{t-1} (1 - c_i) \), where \( c_t \) is the confidence score for the system’s guess at time \( t \).

The model score \( S_t \) for each step is calculated as \( S_t = b_t g_t \), where \( g_t \) is the correctness of the system's guess (\( g_t = 1 \) for a correct answer, \( g_t = 0 \) otherwise). The total system score \( S_q \) for a given question is the sum of scores across all steps: \( S_q = \sum_{t=0}^T S_t \). A key assumption of this framework is that the system must reach full confidence at some point, ensuring \( \sum_{t=0}^T b_t = 1 \). Consequently, if no correct answer is provided earlier, the confidence probability will reach 1 at the final step.

To compare model calibration with human calibration, we introduce the \textbf{human-adjusted calibration score} (\( SH_q \)), which uses human response timing as a benchmark. Human buzz probabilities (\( h_t \)) represent the proportion of humans answering correctly by step \( t \). The score accounts for cases where the system buzzes correctly before humans and when humans do not buzz at all. Let:
\[K_t = h_t \sum_{e=0}^{t} b_e g_e,\]
where \(b_e\) is the system's confidence probability at step \(e\), \(g_e\) indicates whether the system's guess is correct, \(h_t\) is the human buzz probability at step \( t \), and \(e\) is the index representing the step within the range from $0$ to $t$.

The human-adjusted calibration score is then defined as:
\[SH_q = \sum_{t=0}^{T} K_t + \left( 1 - \sum_{t=0}^{T} h_t \right) \sum_{t=0}^{T} b_t g_t,\]
where the first term captures the probability of the system buzzing before humans and being correct, and the second term accounts for cases where humans did not buzz. This adjustment penalizes models for overconfidence in scenarios where humans abstain from answering, ensuring that the calibration score reflects both the model's accuracy and its alignment with human calibration. 

\section{Experiment Details}
We query OpenAI APIs for GPT-4 (gpt-4-0613) and GPT-4o (gpt-4o-2024-08-06) experiments.
For other experiments, we implement model inference with vLLM~\cite{kwon2023efficientmemorymanagementlarge} using Hugging Face model names:
\begin{itemize}
    \item {meta-llama/Meta-Llama-3.1-8B-Instruct}
    \item {meta-llama/Meta-Llama-3.1-70B-Instruct}
    \item {mistralai/Mistral-7B-Instruct-v0.3}
    \item {meta-llama/Llama-2-70b-chat-hf}
\end{itemize}
For 7B/8B models, we use 1 NVIDIA RTX A6000 GPU and 32GB of RAM, and processing each question takes approximately about 5 seconds.
For 70B models, 8 NVIDIA RTX A5000 GPUs and 64GB of RAM are used, and approximately each question takes 40 seconds.
The temperature is set to be 0 for all experiments. The metric computation is highly efficient, taking only 5-10 minutes on a single CPU for datasets of moderate size. Our metric is a single-run metric that evaluates models based on their confidence values. The human-in-the-loop process introduces variability, they are consistent with the cost of crafting other common QA datasets.
We use NLTK, SpaCy, regex, and PEDANTS packages for data pre-processing of the collected buzzpoints.

\section{License Details}
\name{} is licensed under the Creative Commons Attribution 4.0 International License (CC BY 4.0). 
To view a copy of this license, visit \href{https://creativecommons.org/licenses/by/4.0/}{https://creativecommons.org/licenses/by/4.0/} or send a letter to Creative Commons, PO Box 1866, Mountain View, CA 94042, USA.
When using this dataset, please attribute as follows:
\begin{quote}
\name{} is licensed under CC BY 4.0. To view a copy of this license, visit \url{https://creativecommons.org/licenses/by/4.0/}.
\end{quote}

\section{\metric{} Normalization}\label{app:normalization}
For our \metric{}, we apply a normalized sigmoid transformation:
\[\text{\metric{}}(x) = 1 - r \left( \mathbb{E} \left[ (1 - h_t) g_t c_t\right] \right).
\]
$r(x)$ is a normalized sigmoid function designed to map an expected value from a [-1, 1] range to a [0,1] range.
\[r(x) = \frac{\sigma(x) - \sigma(-1)}{\sigma(1) - \sigma(-1)},\]
where
\[\sigma(x) = \frac{1}{1 + e^{-x}}.\]

\section{\metric{} analysis by category}
With logit-based \metric{} scores, GPT-4o performs best in Arts and Science, while LLaMA-2-70B-Chat has the highest error in Arts. GPT-4o also excels in History and Literature, whereas LLaMA-2-70B-Chat struggles most in Literature.

In verbalized-based \metric{} scores, GPT-4o  achieves the lowest calibration errors across Arts and Science. Meanwhile, Mistral-7B-Instruct and LLaMA-2-70B-Chat show the highest errors in Literature, respectively. Verbalized outputs generally show improved calibration compared to logit-based confidence for Mistral and Llama-3 models.
\begin{table*}[]
\resizebox{\textwidth}{!}{
\centering
\begin{tabular}{@{}lcc|cc|cc|cc|cc|cc@{}}
\toprule
\multirow{2}{*}{\textbf{Category}} & \multicolumn{2}{c}{\textbf{GPT-4}} & \multicolumn{2}{c}{\textbf{GPT-4o}} & \multicolumn{2}{c}{\textbf{Mistral-7b-Instruct}} & \multicolumn{2}{c}{\textbf{LLama-2-70b-Chat}} & \multicolumn{2}{c}{\textbf{Llama-3.1-8B-Instruct}} & \multicolumn{2}{c}{\textbf{Llama-3.1-70B-Instruct}} \\
 & \textbf{Logit} & \textbf{Verbalized} & \textbf{Logit} & \textbf{Verbalized} & \textbf{Logit} & \textbf{Verbalized} & \textbf{Logit} & \textbf{Verbalized} & \textbf{Logit} & \textbf{Verbalized} & \textbf{Logit} & \textbf{Verbalized} \\
 \midrule
Arts & 0.6646 & 0.564 & 0.5946 & 0.5592 & 0.7489 & 0.7889 & 0.8063 & 0.7218 & 0.6623 & 0.7466 & 0.6317 & 0.6774 \\
Geo/CE & 0.618 & 0.5465 & 0.6506 & 0.5795 & 0.7193 & 0.7864 & 0.8067 & 0.6872 & 0.6182 & 0.6857 & 0.6307 & 0.6819 \\
History & 0.709 & 0.5874 & 0.691 & 0.6421 & 0.7904 & 0.8022 & 0.8402 & 0.7478 & 0.6844 & 0.803 & 0.6764 & 0.7278 \\
Literature & 0.7315 & 0.628 & 0.7233 & 0.673 & 0.7713 & 0.7773 & 0.8332 & 0.7691 & 0.6861 & 0.7944 & 0.6935 & 0.7162 \\
RMPSS & 0.6676 & 0.5888 & 0.6456 & 0.5812 & 0.8029 & 0.8128 & 0.8514 & 0.755 & 0.6982 & 0.8082 & 0.6669 & 0.7083 \\
Science & 0.5978 & 0.5584 & 0.6019 & 0.5425 & 0.7256 & 0.7691 & 0.792 & 0.7103 & 0.6567 & 0.7325 & 0.5883 & 0.5928 \\
\bottomrule
\end{tabular}
}
\caption{Verbalized and logit-based \metric{} per category. GPT-4 and GPT-4o, the strongest models, struggle with literature and history relative to other categories. Most models are strongest at science. For GPT-4, GPT-4o, and Llama-2-70b, logit-based calibration consistently exhibits higher error; for Mistral and the Llama-3 models, verbalized calibration exhibits higher error.}
\label{tab:calscore-stats}
\end{table*}

%% file: 2025_acl_advcal.bbl
\begin{thebibliography}{38}
\expandafter\ifx\csname natexlab\endcsname\relax\def\natexlab#1{#1}\fi

\bibitem[{Band et~al.(2024)Band, Li, Ma, and Hashimoto}]{band2024linguisticcalibrationlongformgenerations}
Neil Band, Xuechen Li, Tengyu Ma, and Tatsunori Hashimoto. 2024.
\newblock Linguistic calibration of long-form generations.
\newblock In \emph{Proceedings of the 41st International Conference on Machine Learning}, ICML'24. JMLR.org.

\bibitem[{Boyd-Graber et~al.(2012)Boyd-Graber, Satinoff, He, and Daum{\'e}~III}]{boyd-graber-etal-2012-besting}
Jordan Boyd-Graber, Brianna Satinoff, He~He, and Hal Daum{\'e}~III. 2012.
\newblock \href {https://aclanthology.org/D12-1118/} {Besting the quiz master: Crowdsourcing incremental classification games}.
\newblock In \emph{Proceedings of the 2012 Joint Conference on Empirical Methods in Natural Language Processing and Computational Natural Language Learning}, pages 1290--1301, Jeju Island, Korea. Association for Computational Linguistics.

\bibitem[{Brier(1950)}]{brier1950verification}
Glenn~W Brier. 1950.
\newblock Verification of forecasts expressed in terms of probability.
\newblock \emph{Monthly weather review}, 78(1):1--3.

\bibitem[{Caruana(2019)}]{caruana2019friends}
Richard Caruana. 2019.
\newblock \href {https://doi.org/10.1145/3292500.3340414} {Friends don't let friends deploy black-box models: The importance of intelligibility in machine learning}.
\newblock In \emph{Proceedings of the 25th ACM SIGKDD International Conference on Knowledge Discovery \& Data Mining}, KDD '19, page 3174, New York, NY, USA. Association for Computing Machinery.

\bibitem[{Deng et~al.(2025)Deng, Heybati, and Yadav}]{deng2025development}
Jiawen Deng, Kiyan Heybati, and Hemang Yadav. 2025.
\newblock Development and validation of machine-learning models for predicting the risk of hypertriglyceridemia in critically ill patients receiving propofol sedation using retrospective data: a protocol.
\newblock \emph{BMJ open}, 15(1):e092594.

\bibitem[{Engstrom et~al.(2020)Engstrom, Ilyas, Santurkar, Tsipras, Tran, and Madry}]{engstrom2019adversarial}
Logan Engstrom, Andrew Ilyas, Shibani Santurkar, Dimitris Tsipras, Brandon Tran, and Aleksander Madry. 2020.
\newblock Adversarial robustness as a prior for learned representations.
\newblock \emph{arXiv preprint arXiv:1906.00945}.

\bibitem[{Ferrucci(2012)}]{ferrucci2012introduction}
David~A Ferrucci. 2012.
\newblock Introduction to “this is watson”.
\newblock \emph{IBM Journal of Research and Development}, 56(3.4):1--1.

\bibitem[{Guo et~al.(2017)Guo, Pleiss, Sun, and Weinberger}]{guo2017oncalibration}
Chuan Guo, Geoff Pleiss, Yu~Sun, and Kilian~Q. Weinberger. 2017.
\newblock On calibration of modern neural networks.
\newblock In \emph{Proceedings of the 34th International Conference on Machine Learning - Volume 70}, ICML'17, page 1321–1330. JMLR.org.

\bibitem[{He et~al.(2016{\natexlab{a}})He, Boyd-Graber, Kwok, and Daum\'e}]{pmlr-v48-he16}
He~He, Jordan Boyd-Graber, Kevin Kwok, and Hal Daum\'e, III. 2016{\natexlab{a}}.
\newblock \href {https://proceedings.mlr.press/v48/he16.html} {Opponent modeling in deep reinforcement learning}.
\newblock In \emph{Proceedings of The 33rd International Conference on Machine Learning}, volume~48 of \emph{Proceedings of Machine Learning Research}, pages 1804--1813, New York, New York, USA. PMLR.

\bibitem[{He et~al.(2016{\natexlab{b}})He, Boyd-Graber, Kwok, and {Daum\'{e} III}}]{he-16}
He~He, Jordan Boyd-Graber, Kevin Kwok, and Hal {Daum\'{e} III}. 2016{\natexlab{b}}.
\newblock Opponent modeling in deep reinforcement learning.

\bibitem[{Huang et~al.(2023)Huang, Song, Wang, Zhao, Chen, Juefei-Xu, and Ma}]{huang2023look}
Yuheng Huang, Jiayang Song, Zhijie Wang, Shengming Zhao, Huaming Chen, Felix Juefei-Xu, and Lei Ma. 2023.
\newblock Look before you leap: An exploratory study of uncertainty measurement for large language models.
\newblock \emph{arXiv preprint arXiv:2307.10236}.

\bibitem[{Ilia and Aziz(2024)}]{ilia2024predict}
Evgenia Ilia and Wilker Aziz. 2024.
\newblock Predict the next word:< humans exhibit uncertainty in this task and language models \_>.
\newblock In \emph{Proceedings of the 18th Conference of the European Chapter of the Association for Computational Linguistics}, volume~2, pages 234--255.

\bibitem[{Ilyas et~al.(2019)Ilyas, Santurkar, Tsipras, Engstrom, Tran, and Madry}]{ilyas2019adversarial}
Andrew Ilyas, Shibani Santurkar, Dimitris Tsipras, Logan Engstrom, Brandon Tran, and Aleksander Madry. 2019.
\newblock Adversarial examples are not bugs, they are features.
\newblock \emph{Advances in neural information processing systems}, 32.

\bibitem[{Kamath et~al.(2020)Kamath, Jia, and Liang}]{kamath-etal-2020-selective}
Amita Kamath, Robin Jia, and Percy Liang. 2020.
\newblock \href {https://doi.org/10.18653/v1/2020.acl-main.503} {Selective question answering under domain shift}.
\newblock In \emph{Proceedings of the 58th Annual Meeting of the Association for Computational Linguistics}, pages 5684--5696, Online. Association for Computational Linguistics.

\bibitem[{Kaur et~al.(2020)Kaur, Nori, Jenkins, Caruana, Wallach, and Wortman~Vaughan}]{kaur2020interpreting}
Harmanpreet Kaur, Harsha Nori, Samuel Jenkins, Rich Caruana, Hanna Wallach, and Jennifer Wortman~Vaughan. 2020.
\newblock Interpreting interpretability: understanding data scientists' use of interpretability tools for machine learning.
\newblock In \emph{Proceedings of the 2020 CHI conference on human factors in computing systems}, pages 1--14.

\bibitem[{Khurana et~al.(2024)Khurana, Nalisnick, Fokkens, and Swayamdipta}]{khurana2024crowdcalibratorannotatordisagreementinform}
Urja Khurana, Eric Nalisnick, Antske Fokkens, and Swabha Swayamdipta. 2024.
\newblock \href {http://arxiv.org/abs/2408.14141} {Crowd-calibrator: Can annotator disagreement inform calibration in subjective tasks?}

\bibitem[{Kiela et~al.(2021)Kiela, Bartolo, Nie, Kaushik, Geiger, Wu, Vidgen, Prasad, Singh, Ringshia, Ma, Thrush, Riedel, Waseem, Stenetorp, Jia, Bansal, Potts, and Williams}]{kiela-etal-2021-dynabench}
Douwe Kiela, Max Bartolo, Yixin Nie, Divyansh Kaushik, Atticus Geiger, Zhengxuan Wu, Bertie Vidgen, Grusha Prasad, Amanpreet Singh, Pratik Ringshia, Zhiyi Ma, Tristan Thrush, Sebastian Riedel, Zeerak Waseem, Pontus Stenetorp, Robin Jia, Mohit Bansal, Christopher Potts, and Adina Williams. 2021.
\newblock \href {https://doi.org/10.18653/v1/2021.naacl-main.324} {Dynabench: Rethinking benchmarking in {NLP}}.
\newblock In \emph{Proceedings of the 2021 Conference of the North American Chapter of the Association for Computational Linguistics: Human Language Technologies}, pages 4110--4124, Online. Association for Computational Linguistics.

\bibitem[{Krause et~al.(2023)Krause, Tufa, Santamar{\'\i}a, Daza, Khurana, and Vossen}]{krause2023confidently}
Lea Krause, Wondimagegnhue Tufa, Selene~B{\'a}ez Santamar{\'\i}a, Angel Daza, Urja Khurana, and Piek Vossen. 2023.
\newblock Confidently wrong: exploring the calibration and expression of (un) certainty of large language models in a multilingual setting.
\newblock In \emph{Proceedings of the workshop on multimodal, multilingual natural language generation and multilingual WebNLG Challenge (MM-NLG 2023)}, pages 1--9.

\bibitem[{Kwon et~al.(2023)Kwon, Li, Zhuang, Sheng, Zheng, Yu, Gonzalez, Zhang, and Stoica}]{kwon2023efficientmemorymanagementlarge}
Woosuk Kwon, Zhuohan Li, Siyuan Zhuang, Ying Sheng, Lianmin Zheng, Cody~Hao Yu, Joseph~E. Gonzalez, Hao Zhang, and Ion Stoica. 2023.
\newblock \href {http://arxiv.org/abs/2309.06180} {Efficient memory management for large language model serving with pagedattention}.

\bibitem[{Li et~al.(2024)Li, Mondal, Nghiem, Liang, and Boyd-Graber}]{li-etal-2024-pedants}
Zongxia Li, Ishani Mondal, Huy Nghiem, Yijun Liang, and Jordan~Lee Boyd-Graber. 2024.
\newblock \href {https://doi.org/10.18653/v1/2024.findings-emnlp.548} {{PEDANTS}: Cheap but effective and interpretable answer equivalence}.
\newblock In \emph{Findings of the Association for Computational Linguistics: EMNLP 2024}, pages 9373--9398, Miami, Florida, USA. Association for Computational Linguistics.

\bibitem[{Liu et~al.(2024)Liu, Fu, Yogatama, and Neiswanger}]{liu2024dellma}
Ollie Liu, Deqing Fu, Dani Yogatama, and Willie Neiswanger. 2024.
\newblock Dellma: A framework for decision making under uncertainty with large language models.
\newblock \emph{arXiv preprint arXiv:2402.02392}.

\bibitem[{Min et~al.(2020)Min, Michael, Hajishirzi, and Zettlemoyer}]{min2020ambigqa}
Sewon Min, Julian Michael, Hannaneh Hajishirzi, and Luke Zettlemoyer. 2020.
\newblock \href {https://doi.org/10.18653/v1/2020.emnlp-main.466} {{A}mbig{QA}: Answering ambiguous open-domain questions}.
\newblock In \emph{Proceedings of the 2020 Conference on Empirical Methods in Natural Language Processing (EMNLP)}, pages 5783--5797, Online. Association for Computational Linguistics.

\bibitem[{Naeini et~al.(2015)Naeini, Cooper, and Hauskrecht}]{naeini2015obtaining}
Mahdi~Pakdaman Naeini, Gregory Cooper, and Milos Hauskrecht. 2015.
\newblock Obtaining well calibrated probabilities using bayesian binning.
\newblock In \emph{Proceedings of the AAAI conference on artificial intelligence}, volume~29.

\bibitem[{Nguyen and O{'}Connor(2015)}]{nguyen-oconnor-2015-posterior}
Khanh Nguyen and Brendan O{'}Connor. 2015.
\newblock \href {https://doi.org/10.18653/v1/D15-1182} {Posterior calibration and exploratory analysis for natural language processing models}.
\newblock In \emph{Proceedings of the 2015 Conference on Empirical Methods in Natural Language Processing}, pages 1587--1598, Lisbon, Portugal. Association for Computational Linguistics.

\bibitem[{Rodriguez et~al.(2019{\natexlab{a}})Rodriguez, Feng, Iyyer, He, and Boyd-Graber}]{rodriguez2019quizbowl}
Pedro Rodriguez, Shi Feng, Mohit Iyyer, He~He, and Jordan Boyd-Graber. 2019{\natexlab{a}}.
\newblock Quizbowl: The case for incremental question answering.
\newblock \emph{arXiv preprint arXiv:1904.04792}.

\bibitem[{Rodriguez et~al.(2019{\natexlab{b}})Rodriguez, Feng, Iyyer, He, and Boyd{-}Graber}]{quizbowl2019}
Pedro Rodriguez, Shi Feng, Mohit Iyyer, He~He, and Jordan~L. Boyd{-}Graber. 2019{\natexlab{b}}.
\newblock \href {http://arxiv.org/abs/1904.04792} {Quizbowl: The case for incremental question answering}.
\newblock \emph{CoRR}, abs/1904.04792.

\bibitem[{Stengel-Eskin et~al.(2024)Stengel-Eskin, Hase, and Bansal}]{StengelEskin2024LACIELF}
Elias Stengel-Eskin, Peter Hase, and Mohit Bansal. 2024.
\newblock \href {https://api.semanticscholar.org/CorpusID:270199379} {Lacie: Listener-aware finetuning for confidence calibration in large language models}.
\newblock \emph{ArXiv}, abs/2405.21028.

\bibitem[{Stengel-Eskin and Van~Durme(2023)}]{stengel2023calibrated}
Elias Stengel-Eskin and Benjamin Van~Durme. 2023.
\newblock Calibrated interpretation: Confidence estimation in semantic parsing.
\newblock \emph{Transactions of the Association for Computational Linguistics}, 11:1213--1231.

\bibitem[{Tian et~al.(2023)Tian, Mitchell, Zhou, Sharma, Rafailov, Yao, Finn, and Manning}]{tian-etal-2023-just}
Katherine Tian, Eric Mitchell, Allan Zhou, Archit Sharma, Rafael Rafailov, Huaxiu Yao, Chelsea Finn, and Christopher Manning. 2023.
\newblock \href {https://doi.org/10.18653/v1/2023.emnlp-main.330} {Just ask for calibration: Strategies for eliciting calibrated confidence scores from language models fine-tuned with human feedback}.
\newblock In \emph{Proceedings of the 2023 Conference on Empirical Methods in Natural Language Processing}, pages 5433--5442, Singapore. Association for Computational Linguistics.

\bibitem[{Touvron et~al.(2023)Touvron, Martin, Stone, Albert, Almahairi, Babaei, Bashlykov, Batra, Bhargava, Bhosale et~al.}]{touvron2023llama}
Hugo Touvron, Louis Martin, Kevin Stone, Peter Albert, Amjad Almahairi, Yasmine Babaei, Nikolay Bashlykov, Soumya Batra, Prajjwal Bhargava, Shruti Bhosale, et~al. 2023.
\newblock Llama 2: Open foundation and fine-tuned chat models.
\newblock \emph{arXiv preprint arXiv:2307.09288}.

\bibitem[{Ulmer et~al.(2024)Ulmer, Gubri, Lee, Yun, and Oh}]{ulmer-etal-2024-calibrating}
Dennis Ulmer, Martin Gubri, Hwaran Lee, Sangdoo Yun, and Seong Oh. 2024.
\newblock \href {https://doi.org/10.18653/v1/2024.acl-long.824} {Calibrating large language models using their generations only}.
\newblock In \emph{Proceedings of the 62nd Annual Meeting of the Association for Computational Linguistics (Volume 1: Long Papers)}, pages 15440--15459, Bangkok, Thailand. Association for Computational Linguistics.

\bibitem[{Wallace et~al.(2019)Wallace, Rodriguez, Feng, Yamada, and Boyd-Graber}]{wallace2019trick}
Eric Wallace, Pedro Rodriguez, Shi Feng, Ikuya Yamada, and Jordan Boyd-Graber. 2019.
\newblock \href {https://doi.org/10.1162/tacl_a_00279} {Trick me if you can: Human-in-the-loop generation of adversarial examples for question answering}.
\newblock \emph{Transactions of the Association for Computational Linguistics}, 7:387--401.

\bibitem[{Xiong et~al.()Xiong, Hu, Lu, LI, Fu, He, and Hooi}]{xiongcan}
Miao Xiong, Zhiyuan Hu, Xinyang Lu, YIFEI LI, Jie Fu, Junxian He, and Bryan Hooi.
\newblock Can llms express their uncertainty? an empirical evaluation of confidence elicitation in llms.
\newblock In \emph{The Twelfth International Conference on Learning Representations}.

\bibitem[{Xiong et~al.(2024)Xiong, Hu, Lu, Li, Fu, He, and Hooi}]{xiong2024llmsexpressuncertaintyempirical}
Miao Xiong, Zhiyuan Hu, Xinyang Lu, Yifei Li, Jie Fu, Junxian He, and Bryan Hooi. 2024.
\newblock \href {http://arxiv.org/abs/2306.13063} {Can llms express their uncertainty? an empirical evaluation of confidence elicitation in llms}.

\bibitem[{You and Lowd(2022)}]{you-lowd-2022-towards}
Wencong You and Daniel Lowd. 2022.
\newblock \href {https://doi.org/10.18653/v1/2022.nlppower-1.2} {Towards stronger adversarial baselines through human-{AI} collaboration}.
\newblock In \emph{Proceedings of NLP Power! The First Workshop on Efficient Benchmarking in NLP}, pages 11--21, Dublin, Ireland. Association for Computational Linguistics.

\bibitem[{Yu et~al.(2023)Yu, Min, Zettlemoyer, and Hajishirzi}]{yu-etal-2023-crepe}
Xinyan Yu, Sewon Min, Luke Zettlemoyer, and Hannaneh Hajishirzi. 2023.
\newblock \href {https://doi.org/10.18653/v1/2023.acl-long.583} {{CREPE}: Open-domain question answering with false presuppositions}.
\newblock In \emph{Proceedings of the 61st Annual Meeting of the Association for Computational Linguistics (Volume 1: Long Papers)}, pages 10457--10480, Toronto, Canada. Association for Computational Linguistics.

\bibitem[{Zhou et~al.(2024)Zhou, Hwang, Ren, and Sap}]{zhou-etal-2024-relying}
Kaitlyn Zhou, Jena Hwang, Xiang Ren, and Maarten Sap. 2024.
\newblock \href {https://doi.org/10.18653/v1/2024.acl-long.198} {Relying on the unreliable: The impact of language models' reluctance to express uncertainty}.
\newblock In \emph{Proceedings of the 62nd Annual Meeting of the Association for Computational Linguistics (Volume 1: Long Papers)}, pages 3623--3643, Bangkok, Thailand. Association for Computational Linguistics.

\bibitem[{Zhou et~al.(2023)Zhou, Jurafsky, and Hashimoto}]{zhou-etal-2023-navigating}
Kaitlyn Zhou, Dan Jurafsky, and Tatsunori Hashimoto. 2023.
\newblock \href {https://doi.org/10.18653/v1/2023.emnlp-main.335} {Navigating the grey area: How expressions of uncertainty and overconfidence affect language models}.
\newblock In \emph{Proceedings of the 2023 Conference on Empirical Methods in Natural Language Processing}, pages 5506--5524, Singapore. Association for Computational Linguistics.

\end{thebibliography}
